\documentclass[10pt,twocolumn,letterpaper]{article}
\usepackage[final]{cvpr}         % To produce the REVIEW version
\usepackage[accsupp]{axessibility}  % Improves PDF readability for those with disabilities.
\usepackage{graphicx}
\usepackage{amsmath}
\usepackage{amssymb}
\usepackage{booktabs}

\usepackage{comment}
\usepackage{amssymb}
\usepackage{bbm}
\usepackage{array}
\usepackage{multirow}
\usepackage{graphicx}
\usepackage{makecell}

\usepackage{times}
\usepackage{epsfig}
\usepackage{bm}
\usepackage{dsfont}
\usepackage{xcolor}
\usepackage{wrapfig}
\usepackage{color}
\usepackage{verbatim}
\usepackage{caption}
\usepackage{ulem}
\usepackage{placeins}
\usepackage{ragged2e}
\usepackage{diagbox}
\usepackage{enumitem}

\usepackage{textcomp}
\usepackage{soul}

\usepackage{arydshln}
\newcommand{\nickname}{IA-SSD}
\newcommand{\yf}[1]{\textcolor{black}{#1}}
\newcommand{\qy}[1]{\textcolor{black}{#1}}
\newcommand{\todo}[1]{\textcolor{black}{#1}}
\newcommand{\revise}[1]{\textcolor{black}{#1}}

\newcommand{\tabincell}[2]{\begin{tabular}{@{}#1@{}}#2\end{tabular}}

\newcommand{\yfcam}[1]{\textcolor{blue}{#1}}

\usepackage[pagebackref,breaklinks,colorlinks]{hyperref}

\usepackage[capitalize]{cleveref}
\crefname{section}{Sec.}{Secs.}
\Crefname{section}{Section}{Sections}
\Crefname{table}{Table}{Tables}
\crefname{table}{Tab.}{Tabs.}

\def\cvprPaperID{3284} % *** Enter the CVPR Paper ID here
\def\confName{CVPR}
\def\confYear{2022}

\begin{document}
% \title{Not All Points Are Equal: Instance Aware Single-Stage 3D Object Detector}
% \title{\yf{Not All Points Are Equal: Efficient Instance Aware Point-based 3D Object Detector}}
\title{\qy{Not All Points Are Equal: Learning Highly Efficient Point-based Detectors \\ for 3D LiDAR Point Clouds}}

\author{Yifan Zhang\textsuperscript{1}, Qingyong Hu\textsuperscript{2\thanks{Corresponding author}*}, Guoquan Xu\textsuperscript{1},  Yanxin Ma\textsuperscript{1},  Jianwei Wan\textsuperscript{1}, Yulan Guo\textsuperscript{1} \\
\textsuperscript{1}National University of Defense Technology, \textsuperscript{2}University of Oxford \\
\small
\!\!\!\!\!\!\!\!\!\!
\texttt{\{zhangyifan16c, xuguoquan19, mayanxin, wanjianwei, yulan.guo\}@nudt.edu.cn,} 
\hspace{2pt}
\texttt{qingyong.hu@cs.ox.ac.uk}
}
\maketitle

%%%%%%%%% ABSTRACT
\begin{abstract}
We study the problem of efficient object detection of 3D LiDAR point clouds. To reduce the memory and computational cost, existing point-based pipelines usually adopt task-agnostic random sampling or farthest point sampling to progressively downsample input point clouds, despite the fact that not all points are equally important to the task of object detection. In particular, the foreground points are inherently more important than background points for object detectors. Motivated by this, we propose a highly-efficient single-stage point-based 3D detector in this paper, termed \textbf{\nickname{}}. The key of our approach is to exploit two learnable, task-oriented, instance-aware downsampling strategies to hierarchically select the foreground points belonging to objects of interest. Additionally, we also introduce a contextual centroid perception module to further estimate precise instance centers. Finally, we build our \nickname{} following the encoder-only architecture for efficiency. Extensive experiments conducted on several large-scale detection benchmarks demonstrate the competitive performance of our \nickname{}. Thanks to the low memory footprint and a high degree of parallelism, it achieves a superior speed of 80+ frames-per-second on the KITTI dataset with a single RTX2080Ti GPU. The code is available at \url{https://github.com/yifanzhang713/IA-SSD}.
\end{abstract}

%%%%%%%%% BODY TEXT
\section{Introduction}
\label{sec:intro}

\qy{Accurate recognition and localization of specific 3D objects is a fundamental research problem in 3D computer vision \cite{guo2020deep}. As a commonly-used 3D representation, point cloud has attracted increasing attention for its flexibility and compactness. However, the task of 3D object detection in LiDAR point clouds  (\textit{i.e.}, predicting 3D bounding boxes with 7 degrees-of-free including 3D-location, 3D-size, orientation, and class labels) remains highly challenging due to the complex geometrical structure and non-uniform density.} 

% \yf{Current progress on 3D detection mainly towards on improving the accuracy constantly, while the  consumption of computing resources is usually ignored. 
% In this paper, we report the detail efficiency results on different  conditions. 
% In particular, \textit{Parallelism} for memory footprint and \textit{Speed} under different conditions for calculate complexity comparison.}

\begin{figure}[t]
    \begin{center}
    % \vspace{-0.2cm}
        \includegraphics[width=0.5\textwidth]{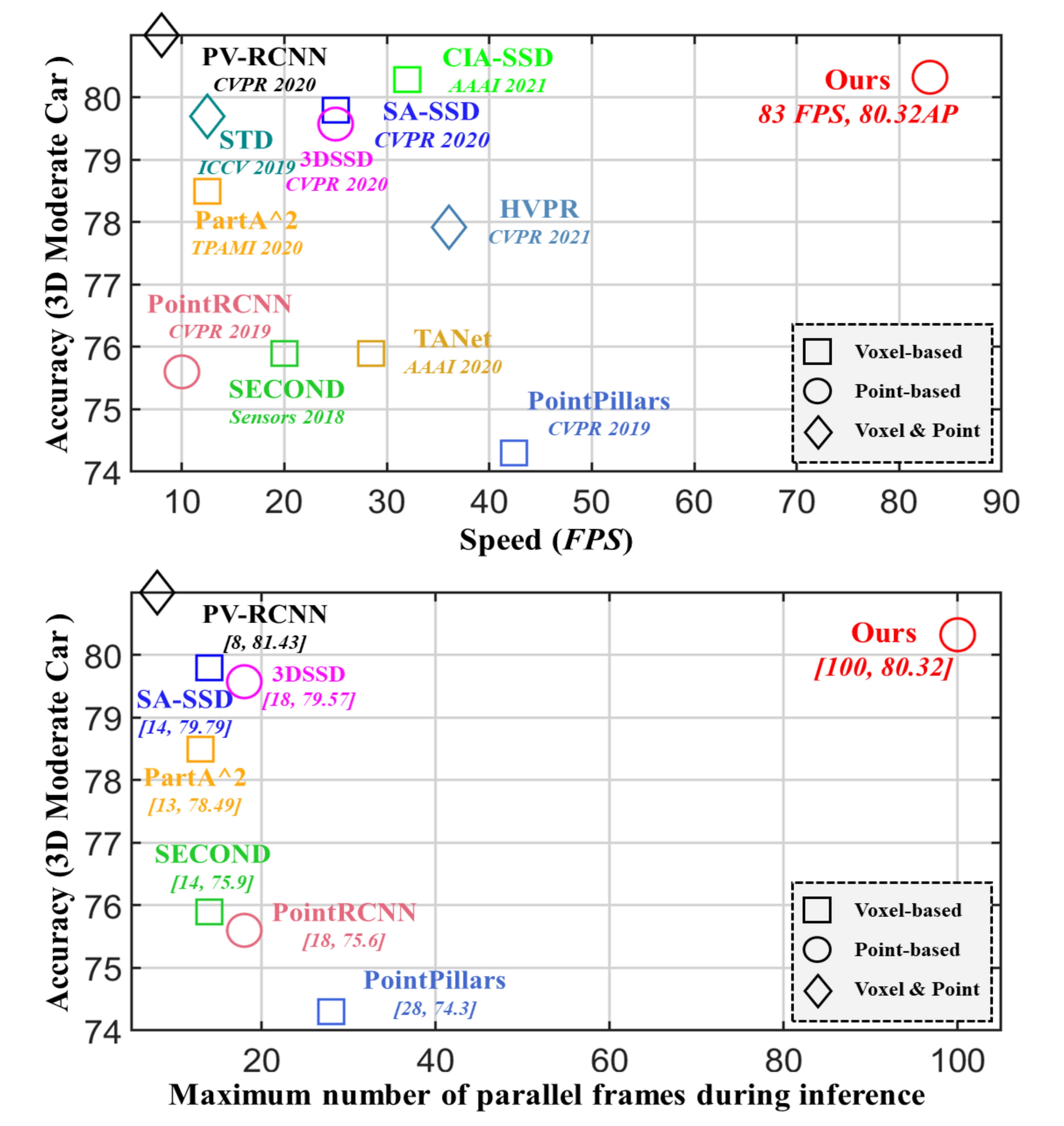}
    \end{center}

   \vspace{-0.3cm}
%   \caption{\qy{The comparison of 3D detection performance and runtime efficiency of different approaches on the KITTI \textit{test} set. In particular, the proposed \nickname{} achieves the top efficiency and comparable performance with the state-of-the-art LiDAR-based detectors. Note that, the STD \cite{yang2019std}, PointRCNN \cite{shi2019pointrcnn}, and Part-$A^2$ \cite{shi2020points} are two-stage detectors, while the rest are single-stage detectors. Best viewed in color.}}
   \caption{\qy{Comparison of the detection performance (accuracy) and efficiency (computational and memory) of different methods in KITTI benchmark.  All experiments are conducted on a single RTX2080Ti GPU. Note that, we evaluate the memory efficiency by calculating the maximum number of parallel frames during inference when fully utilizing the GPU memory. Additionally, the FPS is calculated with the full utilization of GPU memory, more detailed analysis could be found in Table \ref{tab:effi_kitti}.}}
%   by full usage of GPU memory, 
   %detailed comparisons can be found in Table \ref{tab:effi_kitti}}.}}
 %\yf{The comparison of performance with efficiency among different approaches on the KITTI benchmark. Our proposed \nickname{} achieves the top efficiency in both speed and parallelism (memory footprint) with the state-of-the-art LiDAR-based detectors. In particular, \nickname{} makes the point-based stream reintegrated into the state-of-the-arts.}}
   \label{fig1}
   \vspace{-0.3cm}
\end{figure}

\qy{Due to the unstructured and orderless nature of 3D point clouds, early works usually first convert the raw point clouds into intermediate regular representation, including projecting the 3D point clouds into 2D images from birds-eye-view or frontal view \cite{ali2018yolo3d, ku2018joint, lang2019pointpillars, simon2019complexer, simony2018complex, yang2018pixor, zhou2020end}, or transformed into dense 3D voxels \cite{yan2018second, zhou2018voxelnet}. Then, several well-developed 2D detection paradigms can be deployed into the task of 3D object detection. Although remarkable progress has been achieved recently \cite{yang2019std, chen2019fast, liu2020tanet, ye2020hvnet, he2020structure, du2020associate, shi2020points, ao2021spinnet}, these methods introduce the quantization error due to the 3D-2D projection or voxelization, which inevitably limits their performance of existing methods. 
% On the other hand, point-based methods \cite{shi2019pointrcnn, shi2020point, yang20203dssd} 
Another stream of techniques following the point-based pipeline to directly operate on raw point clouds~\
\cite{shi2019pointrcnn, shi2020point, yang20203dssd, hu2021learning, hu2021sqn, hu2022sensaturban, wei2022spatial}. They usually learn point-wise features and then aggregate through specific symmetric functions such as max-pooling \cite{qi2017pointnet, qi2017pointnet++}. Although promising and without any explicit information loss, these methods still suffer from expensive computational/memory costs and limited detection performance.}

%\qy{In this paper, we first dive deep into the existing point-based frameworks and analyze the heuristic sampling strategies used in their pipelines. We then identify that the heuristic sampling strategies used in existing pipelines are far from satisfactory, since most of the important foreground points have been dropped before the bounding box regression step. Fundamentally, the task of 3D object detection is naturally different from other dense prediction tasks such as semantic segmentation, since \textit{not all points are equally important to the task of object detection. }} \qy{In particular, only the foreground points, are the things we really care about.}

\qy{In this paper, we first dive deep into the existing point-based frameworks and experimentally find that the heuristic sampling strategies used are far from satisfactory, since a number of the important foreground points have been dropped before the final bounding box regression step. As such, the detection performance, especially for small objects such as pedestrians, has been fundamentally limited. In this paper, we argue that \textit{not all points are equally important to the task of object detection.}  \qy{In particular, only the foreground points, are the things we really care about.}}

\qy{Motivated by this, we aim to propose a task-oriented, instance-aware downsampling framework, to explicitly preserve foreground points while reducing the memory/computational cost. Specifically, two variants, namely \textit{class-aware} and \textit{centroid-aware} sampling strategies are proposed. In addition, we also present a contextual instance centroid perception, to fully exploit the meaningful context information around bounding boxes for instance center regression. 
% Finally, we build our \nickname{} upon the single-stage framework 3DSSD \cite{yang20203dssd}.
Finally, we build our \nickname{} based on the bottom-up single-stage framework. As shown in Figure \ref{fig1}, the proposed \nickname{} demonstrated to be highly efficient
% (up to 83 frame-per-second (fps) on a single RTX 2080Ti GPU)
\yf{(up to inference 100 frames in parallel in a single pass, with a speed of 83 FPS on a single RTX 2080Ti GPU)}
and accurate on the KITTI benchmark \cite{geiger2012we}. In particular, thanks to the high instance recall ratio of the proposed sampling strategy, the proposed \nickname{} can be directly trained with multiple object categories, rather than the common practice, \textit{i.e.,}  train separate models for different categories. Extensive experiments on Section \ref{sec:experiments} justify the compelling performance and superior efficiency of our method.} 

\qy{To summarize, the contributions are listed as follows:}
\vspace{-0.2cm}

\begin{itemize}
\setlength{\parsep}{0pt} 
\setlength{\topsep}{0pt} 
\setlength{\itemsep}{0pt}
\setlength{\parsep}{0pt}
\setlength{\parskip}{0pt}
    \vspace{-0.1cm}
\item \qy{We identify the sampling issue in existing point-based detectors, and proposed an efficient point-based 3D detector by introducing two learning-based instance-aware downsampling strategies.}
\item \qy{The proposed \nickname{} is highly efficient and capable of detecting multi-class objects on LiDAR point clouds in a single pass. We also provided a detailed memory footprint vs. inference-speed analysis to further validate the superiority of the proposed method.}
\item \qy{Extensive experiments on several large-scale datasets demonstrate the superior efficiency and accurate detection performance of the proposed method.}
\end{itemize}

% \begin{itemize}\vspace{-0.05in}

% \item \qy{We identify the sampling issue in existing point-based detectors, and proposed an efficient point-based 3D detector by introducing two learning-based instance-aware downsampling strategies.}

% \vspace{-0.05in}
% \item \yf{We provide a detailed memory-footprint vs. inference-speed analysis, and show that we can achieve significantly efficient multi-class detection while maintaining the accurate benefits from operating on the raw LiDAR point clouds. }

% % with significant memory and computational friendly over existing baselines\yf{(100-frame parallel)} with the superior processing speed (83 frame-per-second).

% % It is highly efficient and capable of detecting 100-frame scenes in a single pass. Experiment on the KITTI 3D detection benchmark demonstrates 

% \vspace{-0.05in}
% \item \yf{Extensive experiments further demonstrate the superior efficiency and adaptability on more complicated large-scale panoramic scenarios. }
% \end{itemize}

%-------------------------------------------------------------------------
\section{Related Work}
\label{sec:related}
\qy{Here, we give a brief overview of existing voxel-based detectors, point-based detectors, and point-voxel detectors.}

\subsection{Voxel-based Detectors}
\qy{To process unstructured 3D point clouds, voxel-based detectors usually first convert the irregular point clouds into regular voxel grids. This further allows leveraging the mature convolution neural architectures. Early works such as \cite{wang2015voting} densely voxelized the input point clouds and then utilized convolutional neural networks to learn specific geometrical patterns. However, efficiency is one of the main limitations of these methods, since the computational and memory cost grow cubically with the input resolution. To this end, Yan et al. \cite{yan2018second} present an efficient architecture called SECOND by leveraging the 3D submanifold sparse convolution \cite{graham2017submanifold}. By reducing the calculation on empty voxels, the computational and memory efficiency have been significantly improved. Further, PointPillars \cite{lang2019pointpillars} is proposed to further simplify the voxels to pillars (\textit{i.e.,} only voxelization in the plane). 
}
\qy{The existing approaches can be roughly divided into single-stage \cite{he2020structure, ye2020hvnet, zheng2021cia, zhao2020sess, yi2020segvoxelnet, du2020associate} and two-stage detectors \cite{shi2019pointrcnn, yang2019std, shi2020points, shi2020pv, deng2020voxel, shi2021pv}. Albeit simple and efficient, they usually failed to achieve satisfactory detection performance due to the downscaled spatial resolution and insufficient structural information, especially for small objects with sparse points. To this end, He et al. \cite{he2020structure} present SA-SSD to leverage the structure information by introducing an auxiliary network. Ye et al. \cite{ye2020hvnet} introduce a Hybrid Voxel network (HVNet) to attentively aggregate and project the multi-scale feature maps to achieve better performance. Zheng et al. \cite{zheng2021cia} present the Confident IoU-Aware (CIA-SSD) network to extract spatial-semantic features for object detection. In comparison, two-stage detectors can achieve better performance, but with high computational/memory cost. Shi et al. \cite{shi2020points} propose a two-stage detector namely Part-$A^2$, which is composed of the part-aware and aggregation module to exploit the intra-object part locations. Deng et al. \cite{deng2021voxel} extend the PV-RCNN \cite{shi2020pv} by introducing a fully convolutional network to further exploit volumetric representation for raw point cloud and refinement simultaneously.
}

\qy{Overall, voxel-based methods can achieve good detection performance with promising efficiency. However, voxelization inevitably introduces quantization loss. In order to compensate for the structural distortion in the pre-processing phase, complex module design needs to be introduced in~\cite{li2021lidar,mao2021voxel,mao2021pyramid,sheng2021improving,miao2021pvgnet}, which in turn greatly deteriorate the final detection efficiency. Additionally, it is not easy to determine the optimal resolution in practice, considering the complex geometry and various different objects.}
%Although these voxel-based methods are widely used and are able to achieve suprising high inference speed, their limitation can never be neglected: 1) Fine-grained 3D geometry information is lost when voxelization; 2) The size of voxels is the main factor to the detection performance. Recently, many works improve this problem by fusing with other representations, which will be intruced later.

%\noindent{\textbf{Point-based Methods.} Point-based methods directly processing the point cloud and generate proposals from raw point cloud. PointNet \cite{qi2017pointnet} and its variants \cite{qi2017pointnet++,wang2019dynamic,thakur2020dynamic,liu2019relation,wang2019pseudo,qian2020end} could take the raw points as input by using symmetric operation to address the unorderness of point cloud. Based on these advances, PointRCNN \cite{shi2019pointrcnn} proposes a two-stage 3D Region proposal network. At the first stage, proposals are generated based on the segmented foreground points via point-based network backbone, then local information of proposals is appled to regress the accurate bounding box. VoteNet \cite{qi2019deep} reformulates the Hough voting in objects centroids prediction through a deep-learning manner and proposes a new 3D proposal generation mechanism. 3DSSD \cite{yang20203dssd} introduces a fusion sampling strategy comprising the Farthest Point Sampling (FPS) on feature and Euclidean space. PointGNN \cite{shi2020point} presents a graph neural network to address the object detection in point cloud.}

\subsection{Point-based Detectors}

\qy{Different from voxel-based methods, point-based methods \cite{yang20203dssd, shi2019pointrcnn, qi2019deep} directly learning geometry from unstructured point clouds, further generate specific proposals for objects of interest. Considering the orderless nature of 3D point clouds, these methods typically adopt PointNet \cite{qi2017pointnet} and its variants \cite{qi2017pointnet++, wang2019dynamic, thakur2020dynamic, liu2019relation, qian2020end} to aggregate independent point-wise features using symmetric functions. Shi et al. \cite{shi2019pointrcnn} propose PointRCNN, a two-stage 3D region proposal framework for 3D object detection. This method first generates object proposals from segmented foreground points, and high-quality 3D bounding boxes are then regressed by exploiting the semantic feature and local spatial cues.}
Qi et al. \cite{qi2019deep} introduce VoteNet, a one-stage point-based 3D detector based on deep Hough voting to predict the instance centroid. \qy{Inspired by single-stage detectors \cite{liu2016ssd} in 2D images, Yang et al. \cite{yang20203dssd} presents a 3D Single-Stage Detection (3DSSD) framework, while the key is a fusion sampling strategy comprising the Farthest Point Sampling on feature and Euclidean space. PointGNN \cite{shi2020point} is a framework by generalizing graph neural network to 3D object detection.}

\qy{Point-based methods directly operate on the raw point clouds, without any extra preprocessing steps such as voxelization, hence usually intuitive and straightforward. However, the main bottleneck of point-based methods is insufficient learning capacity and limited efficiency.}

\subsection{Point-Voxel Methods}
\qy{To overcome the drawbacks of both point-based methods (\textit{i.e.}, irregular and sparse data access, poor memory locality \cite{point-voxel}) and voxel-based methods (\textit{i.e.}, quantization loss), several methods \cite{yang2019std, chen2019fast, shi2020pv, shi2021pv, jiangty2021vicnet} have started to learning from 3D point clouds using point-voxel joint representations. Specifically, PV-RCNN \cite{shi2020pv} and its follow-up work \cite{shi2021pv} extract point-wise features from voxel abstraction networks to refine the proposals generated from 3D voxel backbone. Further, HVPR \cite{noh2021hvpr}, a single-stage 3D detector, introduces an efficient memory module to augment point-based features, thereby providing a better compromise between accuracy and efficiency. Qian et al. \cite{qian2021boundary} propose a lightweight region aggregation refine network (BANet) via local neighborhood graph construction, which produces more accurate box boundary prediction.
}

Overall, different detection pipelines have their own merits. In this paper, we propose \nickname{}, a single-stage point-based detector, to simultaneously improve the detection accuracy and runtime efficiency. In particular, the key differences between our \nickname{} and existing point-based techniques lie in the instance-aware sampling strategies and the contextual instance centroid perception module, as illustrated in the following sections.

\section{The Proposed \nickname{}}
\label{sec:ia-ssd}

\subsection{Overview}
\label{subsec:overview}

\qy{Different from dense prediction tasks such as 3D semantic segmentation, where point-wise prediction is required, \textit{3D object detection naturally focus on the small yet important foreground objects} (\textit{i.e.}, instances of interest including \textit{car}, \textit{pedestrian}, \textit{etc}.). However, existing point-based detectors usually adopt task-agnostic downsampling approaches such as random sampling \cite{hu2020randla} or farthest point sampling \cite{qi2017pointnet++, yang20203dssd} in their framework. Albeit effective for memory/computational cost reduction, the most important foreground points are also diminished in progressive downsampling. Additionally, due to the large difference in size and geometrical shape of different objects, existing detectors usually train separate models with various carefully tuned hyperparameters for different types of objects. However, this inevitably affects the deployment of these models in practice. Therefore, the objective of this paper is: \textit{Can we train a single point-based model, which is efficient and capable of detecting multi-class objects in a single pass?}}

\begin{figure*}[thb]
    \begin{center}
        \includegraphics[width=1.0\textwidth]{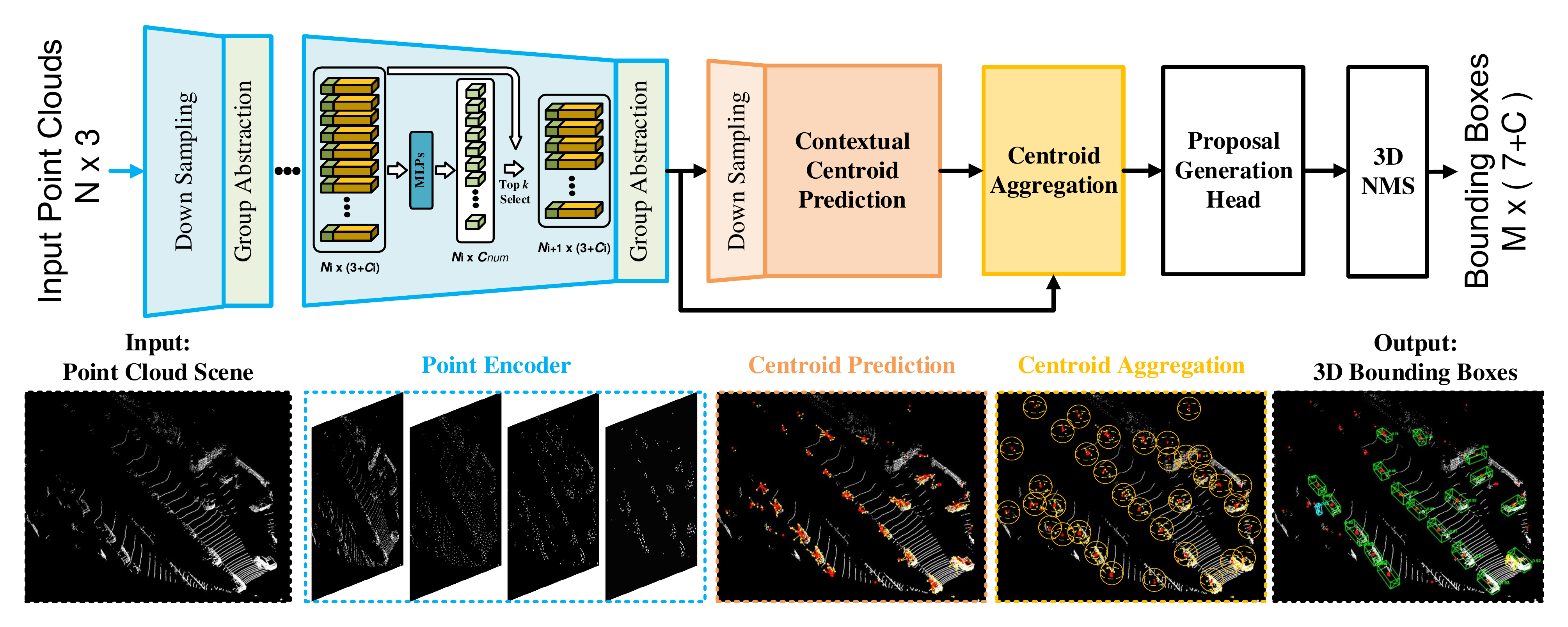}
    \end{center}
%   \centering
%   \includegraphics[width=0.9\textwidth]{figs/fig2_new2.pdf}
  \vspace{-0.4cm}
   \caption{\qy{Illustration of the proposed \nickname{}. The input point clouds are first fed into several Set Abstraction (SA) layers, followed by the instance-aware downsampling to progressively reduce the memory and computational cost. The preserved representative points are further fed into the contextual centroid perception module for instance center prediction and proposal generation. Finally, the 3D bounding box and associated class labels are outputted.}}
   %\caption{Illustration of the \nickname{}. It consists of three main parts. (1) Feature Abstraction layers: Point-wise feature set abstraction with learning-based sampling. (2) Centroid Rally: Points rally to the corresponding instance centroid. (3) Box Generation:  Generating 3D bounding box based on the predicted rally points.} 
   \label{fig2}
  \vspace{-0.4cm}
\end{figure*}

\qy{Motivated by this, we propose an efficient, single-stage detector by introducing the instance-aware downsampling and contextual centroid perception module. As shown in Figure \ref{fig2}, our \nickname{} follows the lightweight encoder-only architecture used in \cite{yang20203dssd} for efficiency. The input LiDAR point clouds are first fed into the network to extract point-wise features, followed by the proposed instance-aware downsampling to progressively reduce the computational cost, while preserving the informative foreground points simultaneously. The learned latent features are further input to the contextual centroid perception module to generate instance proposals and regress the final bounding boxes. 
}

\subsection{Instance-aware Downsampling Strategy}
\label{subsec:instance-aware downsampling strategy}
\qy{For efficient 3D object detection, it is essential to reduce the memory and computational cost through progressive downsampling, especially for large-scale 3D point clouds. However, aggressive downsampling may lose most of the information of the foreground objects. Overall, it remains unclear how to achieve a desirable trade-off between computational efficiency and the preservation of foreground points. To this end, we first conduct an empirical study to quantitatively evaluate different sampling approaches. In particular, we follow the commonly-used encoding architecture (\textit{i.e.,} PointNet++ \cite{qi2017pointnet++} with 4 encoding layers), and report the Instance Recall (\textit{i.e.,} the ratio of instance retained after sampling) at each layer in Table \ref{tab:ins_recall}. In particular, random point sampling \cite{hu2020randla}, FPS based on Euclidean distance (D-FPS) \cite{qi2017pointnet++} and feature distance (Feat-FPS) \cite{yang20203dssd} are reported.}

%In order to further study the capacity of the sampling method for instance preservation, we use Instance Recall - the ratio of the number of instances whose superficial points (also called instance points) preserved after sampling, to help measure the effectiveness of different sampling methods. Different sampling methods and their preservation effects on different classes are listed in Table~\ref{tab1}.

% \begin{figure}[th]
%   \centering
%   \includegraphics[width=0.8\textwidth]{figs/fig3_all2x3.pdf}
%   \vspace{-0.4cm}
%   \caption{Instance recall Among Different Downsampling Strategies, including Rand $\times$ 4, DFS $\times$ 4, Feat-DFS $\times$ 3, Cls-aware $\times$ 3, Cls-aware $\times$ 2 and Ctr-aware $\times$ 2. Note that the DFS is the default sampling method use in the 4096 and 1024 points sampling unless explicitly specified. } 
%   \label{fig3}
%   \vspace{-0.2cm}
% \end{figure}

\textbf{Analysis. }\qy{ It can be seen that: 1) The instance recall rate dropped significantly after several random downsampling operations, indicating massive foreground points have been dropped. 2) Both D-FPS and Feat-FPS achieve a relatively better instance recall rate at the early stage, but also fail to preserve enough foreground points at the last encoding layer. As such, it remains challenging to precisely detect the objects of interest, especially for small objects such as pedestrians and cyclists, where only extremely limited foreground points are left.}

\textbf{Solutions.} \qy{To preserve foreground points as much as possible, we turn to leverage the latent semantics of each point, since the learned point features may incorporate richer semantic information as the hierarchical aggregation operates in each layer. Following this idea, we propose the following two task-oriented sampling approaches by incorporating the foreground semantic priors into the network training pipelines. }

{\bf \textit{Class-aware Sampling}.} \qy{This sampling strategy aims to learn semantics for each point, so as to achieve selective downsampling. To achieve this, we introduce extra branches to exploit the rich semantics in latent features. In particular, two MLP layers were appended to the encoding layers to further estimate the semantic categories of each point. The point-wise one-hot semantic labels generated from the original bounding box annotations are used for supervision. Here we use the vanilla cross-entropy loss:
}

%In order to exploit the semantic feature in SA layer and further lighten the computational burden, we still use semantic feature captured from current SA layer and set two extra MLP layers with Batch Normalization and ReLU activation function to generate a class-aware score for each point to indicate the degree to which the class belong to. The entire forecasting process is trained by introducing GT labels for the supervised learning. To well remit the imbalance issues in classification, we apply the CE as our class-aware sampling loss:

\vspace{-0.2cm}
\begin{footnotesize}
\begin{equation}
   L_{cls\textsc{-}aware}=-\sum_{c=1}^{C}(s_{i}log(\hat{s_{i}})+(1-s_{i})log(1-\hat{s_{i}})) 
%   \label{eql1}
\end{equation}
\end{footnotesize}

\noindent where $C$ denotes the number of categories, $s_{i}$ is the one-hot labels and $\hat{s_{i}}$ denotes the predicted logits. During inference, the points with the top $k$ foreground scores are retained and regarded as the representative points that feed into the next encoding layers. As shown in Table \ref{tab:ins_recall}, this strategy tends to preserve more foreground points, hence achieving a high ratio of instance recall.

%Where $C$ denotes the number of classes, $s_{i}$ donates the binary label of to indicate whether a point falls into a ground-truth box. $s_{i}$ assigned to 1 if the ground truth belongs to the $i$ class and 0 otherwise. Let the probability predicted $s_{t}$ is the output of the MLP layers. After prediction, the top $k$ predicted scores with the points are selected as the reprentative points. In this way, we can achieve points sampling embedded in SA layer with a learning-based manner. 

% Note that during inference, instance points can be remained accompanied with the discriminative power of the SA layers. 

\begin{table*}[t]
   \begin{center}
   \resizebox{0.93\textwidth}{!}{   
   \begin{tabular}{r|c c c|c c c|c c c|c c c}
   \Xhline{2.0\arrayrulewidth}
   \multirow{2}{*}{Sampling strategies}& \multicolumn{3}{c|}{4096 points}& \multicolumn{3}{c|}{1024 points}& \multicolumn{3}{c|}{512 points}& \multicolumn{3}{c}{256 points}  \\
   & \textit{Car} & \textit{Ped.} & \textit{Cyc.} & \textit{Car} & \textit{Ped.} & \textit{Cyc.} & \textit{Car} & \textit{Ped.} & \textit{Cyc.} & \textit{Car} & \textit{Ped.} & \textit{Cyc.} \\
   \Xhline{2.0\arrayrulewidth}
   Random \cite{hu2020randla}      & 96.6$\%$ & 99.1$\%$ & 97.4$\%$ & 87.5$\%$ & 92.7$\%$ & 84.1$\%$ & 78.8$\%$ & 84.9$\%$ & 73.3$\%$ & 67.4$\%$ & 72.1$\%$ & 57.3$\%$  \\
   D-FPS \cite{qi2017pointnet++} & 98.3$\%$ & 100$\%$ & 97.2$\%$ & 97.9$\%$ & 99.3$\%$ & 97.2$\%$ & 96.8$\%$ & 90.6$\%$ & 90.8$\%$ & 91.4$\%$ & 69.1$\%$ & 71.6$\%$  \\
   Feat-FPS \cite{yang20203dssd} & 98.3$\%$ & 100$\%$ & 97.2$\%$ & 97.7$\%$ & 98.0$\%$ & 97.2$\%$ & 96.3$\%$ & 87.6$\%$ & 94.5$\%$ & 95.3$\%$ & 80.1$\%$ & 91.7$\%$  \\
   \textbf{Cls-aware (Ours)} & 98.3$\%$ & 100$\%$ & 97.2$\%$ & 97.9$\%$ & 99.3$\%$ & 97.2$\%$ & 97.9$\%$ & 99.0$\%$ & 95.4$\%$ & 97.9$\%$ & 97.4$\%$ & 92.7$\%$  \\
   \textbf{Ctr-aware (Ours)} & 98.3$\%$ & 100$\%$ & 97.2$\%$ & 97.9$\%$ & 99.3$\%$ & 97.2$\%$ & 97.9$\%$ & 99.0$\%$ & 97.2$\%$ & 97.9$\%$ & 98.4$\%$ & 97.2$\%$  \\
   \Xhline{2.0\arrayrulewidth}
   \end{tabular}}
    \end{center}
    \vspace{-0.2cm}
    \caption{\qy{The instance recall rate for foreground points (\textit{i.e.}, \textit{car}, \textit{pedestrian}, and \textit{cyclist}) after several downsampling on the \revise{entire \textit{validation} set (3769 frames)} of the KITTI benchmark. Note that, the input point clouds with 16384 points are progressively downsampled to 256 points through four downsampling layers. D-FPS are used in the first two layers for the proposed instance aware downsampling strategies.}}
   \label{tab:ins_recall}
   \vspace{-0.1cm}
   \end{table*}
{\bf \textit{Centroid-aware Sampling}.} \revise{Considering instance center estimation is the key for final object detection, we further propose a centroid-aware downsampling strategy to give higher weight to points closer to instance centroid. Specifically, we define the soft point mask of instance $i$ as follows:}

\vspace{-0.2cm}
\begin{footnotesize} 
\begin{equation} 
% Mask_{centerness\raisebox{0mm}{-}3d} = 
\todo{
Mask_{i} = 
\sqrt[3]{\frac{min(f^*, b^*)}{max(f^*, b^*)} 
\times \frac{min(l^*, r^*)}{max(l^*, r^*)} 
\times \frac{min(u^*, d^*)}{max(u^*, d^*)}} 
}
% \label{eql2}
\end{equation} 
\end{footnotesize} 
\vspace{-0.1cm}

\revise{where $f^*, b^*, l^*, r^*, u^*, d^* $ represent the distance of a point to the 6 surfaces (front, back, left, right, up and down) of the bounding box, respectively.  In this case, the point closer to the centroid of the box is likely to have a higher mask score (max value is 1), while the point that lies on the surface will have a mask score of 0. During training, the soft point mask will be used to assign different weights for points within a bounding box based on the spatial locations, hence implicitly incorporates the geometry priors into the network training. In particular, the weighted cross-entropy loss is calculated as follows:}

\begin{footnotesize}
\begin{equation}
   L_{ctr\textsc{-}aware}=-\sum_{c=1}^{C}(Mask_{i} \cdot s_{i}log(\hat{s_{i}})+(1-s_{i})log(1-\hat{s_{i}})) % \label{eql3}
   % L_{cls\raisebox{0mm}{-}aware}=
   % \frac{1}{M_{pos}} \sum_{i}^N -\alpha (1-\hat{s_{t}}^*)^ \gamma \text{log}(\hat{s_{t}}^*)
\end{equation}
\end{footnotesize}

\qy{The soft point mask is multiplied with the loss term of foreground points, so as to assign a higher probability to the points near the center. \revise{Note that, the bounding boxes are no longer required during inference, we simply preserve the top $k$ points with the highest scores after downsampling, if the model is well-trained.}
}

% Based on above setting, with the powerful feature mapping potential, we can sample the instance points that close to the centroid by simply selecting top k predicted scores. 

\subsection{Contextual Instance Centroid Perception}
\label{subsec: contextual instance centroid perception}

% \qy{Predicting accurate instance centroid is a key prerequisite for 3D object detection. However, it remains challenging due to: 1) a 3D object centroid can be far from any surface point \cite{qi2019deep}, hence not easy to localize, especially for sparse and limited foreground points 2) The instance size of different categories vary greatly, thereby non-trivial to use a single network to predict centers for multi-class objects. To this end, we propose a contextual instance centroid perception module to exploit the wide context around the bounding boxes.
% }

\textbf{Contextual Centroid Prediction.} \qy{Inspired by the success of context prediction in 2D images \cite{yang2008context, divvala2009empirical}, we attempt to leverage the contextual cues around the bounding box for instance centroid prediction. \revise{Specifically, we follow \cite{qi2019deep} to explicitly predict an offset $\Delta \hat{c_{ij}}$ to the instance center. Additionally, a regularization term is added to minimize the uncertainty of the centroid prediction. Specifically, all votes per instance are aggregated in light of the surrounding interference, where the $\overline{c_{i}}$ is the mean destination of  $i$-th instance. Therefore, the centroid prediction loss is formulated as follows:}}

\begin{scriptsize}
\begin{equation}
\begin{aligned}
 L_{cent} = &
 \yf{
   \frac{1}{|\mathcal{F_{+}}|} 
   \frac{1}{|\mathcal{S_{+}}|} 
   \sum_{i}
   \sum_{j} (|\Delta \hat{c_{ij}} - \Delta c_{ij}| + | \hat{c_{ij}} - \overline{c_{i}}|)
   \cdot
  \mathbf{I}_{\mathcal{S}}(p_{ij})} \\ 
 &\yf{ where \quad \overline{c_{i}} = \frac{1}{|\mathcal{S_{+}}|} \sum_{j}\hat{c_{ij}}, \quad \mathbf{I}_{\mathcal{S}}:\mathcal{P}\rightarrow \left \{ 0,1 \right \}}
\end{aligned}
%   \label{eql4}
\end{equation}
\end{scriptsize}
\vspace{-0.1cm}

\revise{where $\Delta c_{ij}$ denotes the ground-truth offset from point $p_{ij}$ to the center point. $\mathbf{I}_{\mathcal{S}}$ is an indicator function to determine whether this point is used to estimate the instance center or not.  $|S_{+}|$ is the number of points used to predict the instance center. Note that,  instead of only using the points or the shifted points within the bounding box for instance center prediction \cite{qi2019deep, yang20203dssd}, we also exploit the surrounding representative points from a large context for centroid prediction in this paper. Specifically, we empirically investigate the impact of simple contextual cues on final detection performance. In particular, we manually expand the ground-truth bounding boxes or proportional enlarge the box to cover more related context near the objects. The sampled points that fall in the expanded bounding box are utilized to estimate offset and then shifted.}

\textbf{Centroid-based Instance Aggregation.} \qy{For shifted representative (centroid) points, we further utilize a PointNet++ module to learn a latent representation for each instance. Specifically, we transform the neighboring points to a local canonical coordinate system, then aggregate the point feature through shared MLPs and symmetric functions. }

\textbf{Proposal Generation Head.} \qy{The aggregated centroid point features are then fed into proposal generation head to predict bounding boxes with classes. We encode the proposal as a multidimensional representation with location, scale, and orientation. Finally, all proposals are filtered by 3D-NMS post-processing 
% with an IoU threshold of 0.01.} 
with a specific IoU threshold.}

\subsection{End-to-End Learning}
\label{subsec: end-to-end learning}
\qy{Our \nickname{} can be trained in an end-to-end fashion. Multi-task loss is used in our framework for joint optimization. The total loss $L_{total}$ is composed of downsampling strategy loss $L_{sample}$, centroid prediction loss $L_{cent}$, classification loss $L_{cls}$ and box generation loss $L_{box}$:
}

\vspace{-0.1cm}
\begin{small}
\begin{equation}
   L_{total} = L_{sample} + L_{cent} + L_{cls} + L_{box}  
\label{eql5}
\end{equation}
\end{small}
% \vspace{-0.1cm}

\qy{In particular, the box generation loss can be further decomposed into location, size, angle-bin, angle-res, and corner parts: }

%\yf{For proposal generation, we apply cross-entropy loss for multi-class regulation with IoU mask for all predicted centroid points as $L_{cls}$, and decompose the box generation loss into location, size, angle-bin, angle-res, and corner parts: }

\vspace{-0.2cm}
\begin{small}
\begin{equation}
   L_{box} = L_{loc} + L_{size} + L_{angle\raisebox{0mm}{-}bin} + L_{angle\raisebox{0mm}{-}res} + L_{corner}
\label{eql6}
\end{equation}
\end{small}
\vspace{-0.1cm}
% We apply the ${smooth\textsc{-}l_{1}}$ loss for $L_{loc}$, $L_{size}$, $L_{size}$, $L_{corner}$ and $L_{angle-res}$.

%-------------------------------------------------------------
\section{Experiments}
\label{sec:experiments}

\subsection{Implementation Details}
\qy{
\qy{We build our \nickname{} based on single-stage, encoder-only architecture for efficiency. Specifically, a number of SA layers~\cite{qi2017pointnet++} are used to extract point-wise features.} Multi-scale grouping with increasing radius groups is used ([0.2, 0.8], [0.8, 1.6], [1.6, 4.8]) to steady extract local geometrical features. Considering limited semantics incorporated in early layers, we adopt D-FPS in the first two encoding layers, followed by the proposed instance-aware downsampling. Next, 256 representative point features are fed into the contextual centroid prediction module, followed by three MLP layers (256$\rightarrow$256$\rightarrow$3) to predict the instance centroid. Finally, the classification and regression layers (three MLP layers) are appended to output the semantic labels and the corresponding bounding boxes. More implementation details are reported in the Appendix.
}

\begin{table*}[t]
   \begin{center}
   \resizebox{\textwidth}{!}{  
   \begin{tabular}{c|r|r|c||c c c|c c c|c c c|c}
\Xhline{2.0\arrayrulewidth}
   \multirow{2}{*}{}& \multirow{2}{*}{Method}& \multirow{2}{*}{Reference}& \multirow{2}{*}{Type}& \multicolumn{3}{c|}{3D Car\quad(IoU=0.7)}& \multicolumn{3}{c|}{3D Ped. (IoU=0.5)}& \multicolumn{3}{c|}{3D Cyc. (IoU=0.5)}&   \multirow{2}{*}{Speed} \\
   & & & & \textit{Easy} & \textit{Mod.} & \textit{Hard} & \textit{Easy} & \textit{Mod.} & \textit{Hard} & \textit{Easy} & \textit{Mod.} & \textit{Hard}  \\
\Xhline{2.0\arrayrulewidth}

   \multirow{8}{*}{\rotatebox{90}{Voxel-based}}
   & VoxelNet \cite{zhou2018voxelnet} & CVPR 2018 & 1-stage & 77.47 & 65.11 & 57.73 & 39.48 & 33.69 & 31.5 & 61.22 & 48.36 & 44.37 & 4.5 \\ 
   & SECOND \cite{yan2018second} & Sensors 2018 & 1-stage & 84.65 & 75.96 & 68.71 & 45.31 & 35.52 & 33.14 & 75.83 & 60.82 & 53.67 & 20  \\ 
   & PointPillars \cite{lang2019pointpillars} & CVPR 2019 & 1-stage & 82.58 & 74.31 & 68.99 & 51.45 & 41.92 & 38.89 & 77.10 & 58.65 & 51.92 & 42.4 \\  
   & 3D IoU Loss \cite{zhou2019iou} & 3DV 2019 & 1-stage & 86.16 & 76.50 & 71.39 & - & - & - & - & - & - & 12.5 \\
%   & TANet \cite{liu2020tanet} & CVPR 2020 & 1-stage & 84.39 & 75.94 & 68.82 & 53.72 & \underline{44.34} & 40.49 & 75.70 & 59.44 & 52.53 & 28.5 \\   
   & Associate-3Ddet \cite{du2020associate} & CVPR 2020 & 1-stage & 85.99 & 77.40 & 70.53 & - & - & - & - & - & - & 20 \\  
   & SA-SSD \cite{he2020structure} & CVPR 2020 & 1-stage & 88.75 & 79.79 & 74.16 & - & - & - & - & - & - & 25 \\ 
   & CIA-SSD \cite{zheng2021cia} & AAAI 2021 & 1-stage & 89.59 & 80.28 & 72.87 & - & - & - & - & - & - & 32 \\ 
   & TANet \cite{liu2020tanet} & AAAI 2020 & 2-stage & 84.39 & 75.94 & 68.82 & 53.72 & \underline{44.34} & 40.49 & 75.70 & 59.44 & 52.53 & 28.5 \\  
   & Part-$A^2$ \cite{shi2020points} & TPAMI 2020 & 2-stage & 87.81 & 78.49 & 73.51& 53.10  & 43.35 & 40.06 & 79.17 & 63.52 & 56.93 & 12.5 \\
   \hline
   \multirow{5}{*}{\rotatebox{90}{Point-Voxel}}
   & Fast Point R-CNN \cite{chen2019fast} & ICCV 2019 & 2-stage & 85.29 & 77.40 & 70.24 & -  & - & - & - & - & - & 16.7 \\
   & STD \cite{yang2019std} & ICCV 2019 & 2-stage & 87.95 & 79.71 & 75.09 & 53.29 & 42.47 & 38.35 & 78.69 & 61.59 & 55.30 & 12.5 \\
   & PV-RCNN \cite{shi2020pv}  & CVPR 2020 & 2-stage & \underline{90.25} & \underline{81.43} & \underline{76.82} & 52.17 & 43.29 & 40.29 & 78.60 & 63.71 & \underline{57.65} & 12.5 \\ 
   & VIC-Net \cite{jiangty2021vicnet}  & ICRA 2021 & 1-stage & 88.25 & 80.61 & 75.83 & 43.82  & 37.18 & 35.35 & 78.29 & 63.65 & 57.27 & 17 \\ 
   & HVPR \cite{noh2021hvpr} & CVPR 2021 & 1-stage & 86.38 & 77.92 & 73.04 & 53.47 & 43.96 & \underline{40.64} & - & - & - & 36.1 \\ 
   \hline
   \multirow{8}{*}{\rotatebox{90}{Point-based}} 
   & PointRCNN \cite{shi2019pointrcnn} & CVPR 2019 & 2-stage & 86.96 & 75.64 & 70.70 & 47.98 & 39.37 & 36.01 & 74.96 & 58.82 & 52.53 & 10 \\ 
   & 3D IoU-Net \cite{li20203d} & Arxiv 2020 & 2-stage & 87.96 & 79.03 & 72.78 & -  & - & - & - & - & - & 10 \\
   & Point-GNN \cite{shi2020point} & CVPR 2020 & 1-stage & 88.33 & 79.47 & 72.29 & 51.92 & 43.77 & 40.14 & 78.60 & 63.48 & 57.08 & 1.6 \\
   & 3DSSD \cite{yang20203dssd} & CVPR 2020 & 1-stage & 88.36 & 79.57 & 74.55 & \underline{54.64} & 44.27 & 40.23 & \underline{82.48} & 64.10 & 56.90 & 25  \\ 
   & \todo{3DSSD$^\dagger$(Reproduced)} & CVPR 2020 & 1-stage & 87.73 & 78.58 & 72.01 & 35.03 & 27.76 & 26.08 & 66.69 & 59.00 & 55.62 & \yf{23} \\
   & \todo{3DSSD$^\ddagger$(OpenPCDet)} & CVPR 2020 & 1-stage & 87.91 & 79.55 & 74.71 & 3.63 & 3.18 & 2.57 & 27.08 & 21.38 & 19.68 & \yf{28} \\
   \cline{2-14}
   & \todo{\textbf{\nickname{} (single-class)}}  & \textbf{-} & \textbf{1-stage} & \textbf{88.87} & \textbf{80.32} & \textbf{75.10} & \textbf{49.01} & \textbf{41.20} & \textbf{38.03} & \textbf{80.78} & \underline{\textbf{66.01}} & \underline{\textbf{58.12}} & \underline{\textbf{\yf{85}}} \\
   & \todo{\textbf{\nickname{} (multi-class)}}  & \textbf{-} & \textbf{1-stage} & \textbf{88.34} & \textbf{80.13} & \textbf{75.04} & \textbf{46.51} & \textbf{39.03} & \textbf{35.60} & \textbf{78.35} & \textbf{61.94} & \textbf{55.70} & \underline{\textbf{\yf{83}}} \\
\Xhline{2.0\arrayrulewidth}
   \end{tabular}}
   \end{center}
   \vspace{-0.3cm}
   \caption{\qy{Quantitative detection performance achieved by different methods on the KITTI \textit{test} set. All results are evaluated by mean Average Precision with 40 recall positions via the oﬀicial KITTI evaluation server. The results of our \nickname{} are shown in bold, and the best results are underlined.}
    }   
   \label{tab:kitti_test}
   \vspace{-0.5cm}
   \end{table*}

\begin{table}[h]
\centering
% \vspace{0.3cm}
      \resizebox{0.47\textwidth}{!}{   
      \begin{tabular}{c|r|r|c||p{14mm}<{\centering}|p{14mm}<{\centering}|p{14mm}<{\centering}}
      \Xhline{2.0\arrayrulewidth}
      \multirow{2}{*}{} & \multirow{2}{*}{Method} & \multirow{2}{*}{References} & \multirow{2}{*}{Type} & \small{Car~ Mod} & \small{Ped. Mod} & \small{Cyc. Mod}  \\
      % & & & & & & & Easy & Mod. & Hard & Easy & Mod. & Hard & Easy & Mod. & Hard  \\
      & & & & (IoU=0.7) & (IoU=0.5) & (IoU=0.5)  \\
      \Xhline{2.0\arrayrulewidth}
      \multirow{8}{*}{\rotatebox{90}{Voxel-based}}
      & VoxelNet \cite{zhou2018voxelnet} & CVPR 2018 & 1-stage & 65.46 & 53.42 & 47.65  \\
      & SECOND \cite{yan2018second}   & Sensors 2018 & 1-stage & 76.48 & - & -  \\
      & PointPillars \cite{lang2019pointpillars}   & CVPR 2019 & 1-stage & 77.98  & - & -  \\
      & TANet \cite{liu2020tanet}    & AAAI 2020 & 1-stage & 77.85 & 63.45 & 64.95  \\
      %   & HVNet \cite{ye2020hvnet}    & CVPR 2020  & 77.58 & 64.81 & 73.75  \\
      & Associate-3Ddet \cite{du2020associate} & CVPR 2020 & 1-stage & 79.17 & - & -  \\
      & SA-SSD \cite{he2020structure}   & CVPR 2020 & 1-stage & 79.91 & - & -  \\
      & CIA-SSD \cite{zheng2021cia}  & AAAI 2021 & 1-stage & 79.81 & - & -  \\
      & Part-$A^2$ \cite{shi2020points} & TPAMI 2020 & 2-stage & 79.47  & \underline{63.84} & \underline{73.07}  \\
      \hline
      \multirow{4}{*}{\rotatebox{90}{Point-Voxel}}
      & Fast Point R-CNN \cite{chen2019fast} & ICCV 2019 & 2-stage & 79.00 & - & -  \\  
      & STD \cite{yang2019std}      & ICCV 2019 & 2-stage & 79.8 & - & -  \\
      & PV-RCNN \cite{shi2020pv}  & CVPR 2020 & 2-stage & \underline{83.90}  & - & -  \\
      & VIC-Net \cite{jiangty2021vicnet}  & ICRA 2021 & 1-stage  & 79.25  & - & -  \\
      \hline
      \multirow{5}{*}{\rotatebox{90}{Point-based}}
      & PointRCNN \cite{shi2019pointrcnn} & CVPR 2019 & 2-stage & 78.63  & - & -  \\
      & 3D IoU-Net \cite{li20203d}    & Arxiv 2020 & 2-stage & 79.26 & - & -  \\
      & PointGNN \cite{shi2020point}  & CVPR 2020 &  1-stage & 78.34 & - & -  \\
      & 3DSSD \cite{yang20203dssd}    & CVPR 2020 & 1-stage & 79.45 & -  & -  \\
      \cline{2-7}
      & \textbf{\nickname{} (Ours) } & \textbf{-} & \textbf{1-stage} & \textbf{79.57 } & \textbf{58.91} & \textbf{71.24}  \\
    %   & \textbf{\nickname{} Separated} & - & \textbf{ } & \textbf{60.02} & \textbf{71.89}  \\ 
    \Xhline{2.0\arrayrulewidth}
   \end{tabular}}
   \caption{\qy{Quantitative comparison of different approaches on the \textit{validation} split of the KITTI dataset. The average precision is measured with 11 recall positions (vs. 40 recall positions in the KITTI \textit{test} set) \cite{geiger2012we}. The results achieved by our \nickname{} are shown in bold, while the top-performed results are shown in underline.}}
   \label{tab:kitti_val}
  \vspace{-0.2cm}
   \end{table}  

\subsection{Comparison with State-of-the-Art Methods}

\qy{\textbf{Evaluation on KITTI Dataset.} In the KITTI benchmark, objects belong to \textit{car}, \textit{pedestrian} and \textit{cyclist} are classified into three subsets (``Easy'', ``Moderate'' and ``Hard'') based on the levels of difficulty. The results on ``Moderate'' are usually adopted as the main indicator for final ranking. We report the results achieved by different methods (voxel, point, and point-voxel-based methods) on the test set of the KITTI dataset in Table~\ref{tab:kitti_test}. Note that, since \cite{yang20203dssd} does not provide reproducible implementation or pre-trained models for \textit{pedestrian} and \textit{cyclist}, we have no choice but to provide both the results reported in their paper, the best-reproduced results, and the results achieved by OpenPCDet\footnote{https://github.com/open-mmlab/OpenPCDet} %~\cite{openpcdet2020} 
implementation for a fair comparison. }

\begin{table*}[t]
   \begin{center}
   \resizebox{\textwidth}{!}{  
   \begin{tabular}{r|c||c c|c c|c c|c c|c c|c c}
\Xhline{2.0\arrayrulewidth}
   \multirow{2}{*}{Method} & \multirow{2}{*}{Type} & \multicolumn{2}{c|}{Vehicle (LEVEL 1)} &\multicolumn{2}{c|}{Vehicle (LEVEL 2)} & \multicolumn{2}{c|}{Ped. (LEVEL 1)} & \multicolumn{2}{c|}{Ped. (LEVEL 2)} & \multicolumn{2}{c|}{Cyc. (LEVEL 1)} & \multicolumn{2}{c}{Cyc. (LEVEL 2)} \\
   & & mAP & mAPH & mAP & mAPH & mAP & mAPH & mAP & mAPH & mAP & mAPH & mAP & mAPH \\
    % & & \textit{mAP} & \textit{mAPH} & \textit{mAP} & \textit{mAPH} & \textit{mAP} & \textit{mAPH} & \textit{mAP} & \textit{mAPH} & \textit{mAP} & \textit{mAPH} & \textit{mAP} & \textit{mAPH}\\
\Xhline{2.0\arrayrulewidth}
    PointPillars \cite{lang2019pointpillars} & Voxel-based  & 60.67 & 59.79 & 52.78 & 52.01 & 43.49 & 23.51 & 37.32 & 20.17 & 35.94 & 28.34 & 34.60 & 27.29 \\
    SECOND \cite{yan2018second} & Voxel-based  & 68.03 & 67.44 & 59.57 & 59.04 & 61.14 & 50.33 & 53.00 & 43.56 & 54.66 & 53.31 & 52.67 & 51.37 \\ 
    Part-$A^2$ \cite{shi2020points} & Voxel-based  & 71.82 & 71.29 & 64.33 & 63.82 & 63.15 & 54.96 & 54.24 & 47.11 & 65.23 & 63.92 & 62.61 & 61.35 \\ 
    PV-RCNN \cite{shi2020pv} & Point-Voxel   & \underline{74.06} & \underline{73.38} & \underline{64.99} & \underline{64.38} & 62.66 & 52.68 & 53.80 & 45.14 & 63.32 & 61.71 & 60.72 & 59.18 \\ 
    \textbf{IA-SSD (Ours)} & Point-based  & \textbf{70.53} & \textbf{69.67} & \textbf{61.55} & \textbf{60.80} & \underline{\textbf{69.38}} & \underline{\textbf{58.47}} & \underline{\textbf{60.30}} & \underline{\textbf{50.73}} & \underline{\textbf{67.67}} & \underline{\textbf{65.30}} & \underline{\textbf{64.98}} & \underline{\textbf{62.71}} \\ 
\Xhline{2.0\arrayrulewidth}
   \end{tabular}}
   \end{center}
   \vspace{-0.4cm}
   \caption{\todo{Quantitative detection performance achieved by different methods on the Waymo \cite{sun2020scalability} \textit{validation} set. The results of our \nickname{} are shown in bold, and the best results are underlined.}
}   
   \label{tab:waymo_val}
%   \vspace{-0.2cm}
   \end{table*}

\begin{table*}[t]
   \begin{center}
   \resizebox{\textwidth}{!}{  
   \begin{tabular}{r|r||c c c c|c c c c|c c c c|c}
\Xhline{2.0\arrayrulewidth}
   \multirow{2}{*}{Method} & \multirow{2}{*}{Type}  & \multicolumn{4}{c|}{Vehicle} &\multicolumn{4}{c|}{Pedestrian} & \multicolumn{4}{c|}{Cyclist} & \multirow{2}{*}{mAP} \\
   & &  overall & 0-30m & 30-50m & \textgreater 50m & overall & 0-30m & 30-50m & \textgreater 50m & overall & 0-30m & 30-50m & \textgreater 50m &  \\
\Xhline{2.0\arrayrulewidth}
    PointPillars \cite{lang2019pointpillars} & Voxel-based  & 68.57 & 80.86 & 62.07 & 47.04 & 17.63 & 19.74 & 15.15 & 10.23 & 46.81 & 58.33 & 40.32 & 25.86 & 44.34 \\ 
    SECOND \cite{yan2018second} & Voxel-based  & 71.19 & 84.04 & 63.02 & 47.25 & 26.44 & 29.33 & 24.05 & 18.05 & 58.04 & 69.96 & 52.43 & 34.61 & 51.89 \\ 
    CenterPoints \cite{yin2021center} & Voxel-based  & 66.79 & 80.10 & 59.55 & 43.39 &  \underline{49.90} &  \underline{56.24} &  \underline{42.61} &  \underline{26.27} & \underline{63.45} &  \underline{74.28} & \underline{57.94} & \underline{41.48} & \underline{60.05} \\ 
    PV-RCNN \cite{shi2020pv} & Point-Voxel   & \underline{77.77} & \underline{89.39} & \underline{72.55} & \underline{58.64} & 23.50 & 25.61 & 22.84 & 17.27 & 59.37 & 71.66 & 52.58 & 36.17 & 53.55 \\ 
    PointRCNN \cite{shi2019pointrcnn} & Point-based  & 52.09 & 74.45 & 40.89 & 16.81 & 4.28 & 6.17 & 2.40 & 0.91 & 29.84 & 46.03 & 20.94 & 5.46 & 28.74 \\
    \textbf{IA-SSD (Ours)} & Point-based  & \textbf{70.30} & \textbf{83.01} & \textbf{62.84} & \textbf{47.01} & \textbf{39.82} & \textbf{47.45} & \textbf{32.75} & \textbf{18.99} & \textbf{62.17} & \textbf{73.78} & \textbf{56.31} & \textbf{39.53} & \textbf{57.43}  \\ 
\Xhline{2.0\arrayrulewidth}
   \end{tabular}}
   \end{center}
   \vspace{-0.4cm}
   \caption{Quantitative detection performance on the ONCE \cite{mao2021one} \textit{validation} set. The results of our \nickname{} are shown in bold, and the best results are underlined.
}   
   \label{tab:once_val}
   \vspace{-0.3cm}
   \end{table*}

\begin{figure*}[t]
\centering
\includegraphics[width=0.88\textwidth]{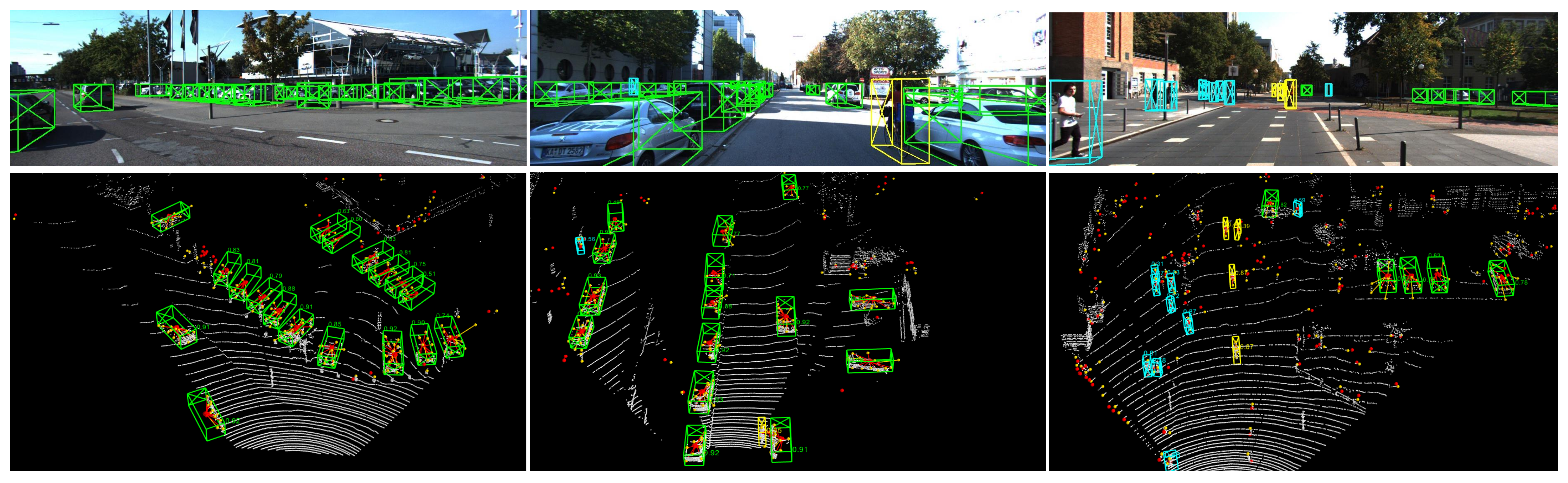}
\vspace{-0.3cm}
\caption{\qy{Qualitative results achieved on the KITTI \textit{test} set. Red point for centroid perception, while gold points denote the 256 representative points. Green boxes for \textit{car}, cyan for \textit{pedestrian} and yellow for \textit{cyclist}. Best viewed in color.}}
\vspace{-0.2cm}
\label{fig3}
\end{figure*}

\textbf{Analysis.} \qy{It can be seen that: 1) the proposed \nickname{} achieves the best \textit{cyclist} detection performance, even outperforming several strong point-voxel and voxel detectors \cite{shi2020points, shi2020pv}. This is mainly because the proposed instance aware sampling can effectively preserve foreground points, enabling accurate detection of small objects. 2) Our \nickname{} also achieves best \textit{car} detection performance compared with other point-based detectors, outperforming PointRCNN \cite{shi2019pointrcnn} by (1.91$\%$, 4.68$\%$, 4.4$\%$), and the SoTA method 3DSSD \cite{yang20203dssd} by (0.51$\%$, 0.75$\%$, 0.55$\%$) mAP.  3) Despite the competitive detection performance, the proposed \nickname{} also shows superior efficiency. It can detect with a speed of 85 FPS on a single NVIDIA RTX 2080Ti with Intel I9-10900X CPU@3.7GHz. 4) Thanks to the instance-aware sampling strategy and the contextual centroid perception module, our framework can be trained with multi-class together (\textit{i.e.}, training a single model for detecting multi-class objects), rather than training separate models for different objects \cite{yang20203dssd}. In particular, the performance is still comparable with other state-of-the-art approaches. This allows our model much more efficient and flexible during inference. Finally, we also show the qualitative results achieved by our \nickname{} in Figure \ref{fig3}. We can clearly see that the proposed \nickname{} 
is capable of detecting small and far-away instances such as \textit{pedestrian} and \textit{cyclist}.
}

% \textbf{Performance comparison on the \textit{validation} split of KITTI dataset. } 
\qy{Apart from the detection results on the \textit{test} split, we also report the performance comparison on the \textit{validation} set of the KITTI dataset in Table \ref{tab:kitti_val}. We can see our \nickname{} achieved the best performance for all three classes among all point-based detectors. In particular, our \nickname{} is single-stage, lightweight, and efficient, requiring only a single model for detecting multi-class objects.}

\revise{\textbf{Evaluation on Waymo Dataset.} We further evaluate the performance of our \nickname{} on Waymo \cite{sun2020scalability} dataset. This dataset is composed of nearly 160k 360-degree LiDAR samples in the \textit{training} set and 40k in the \textit{validation} set with panoramic annotated objects. For a fair comparison, we adapt our framework on the Waymo Dataset by only changing the number of input points from 16384 to 65536, and increasing the sampling scale up to fourfold in each sampling layer, while remaining the rest unchanged. Additionally, all baselines are implemented based on the OpenPCDet codebase for a rigorous comparison. As shown in Table \ref{tab:waymo_val}, our \nickname{} achieves significantly better detection performance on \textit{pedestrian} and \textit{cyclist} compared with other strong baselines, showing that the proposed instance-aware sampling can indeed improve the perception capacity of small objects. We also noticed that our \nickname{} shows slightly inferior detection performance on \textit{vehicle} compared with other voxel-based methods, this may be caused by the relatively complicated distribution of the 3D size of such instances. We will leave this issue for future exploration.}

\qy{\textbf{Evaluation on ONCE Dataset.} To further verify the generalization of \nickname{} on more complex and realistic situations, we also evaluate the performance of our \nickname{} on the latest ONCE Dataset~\cite{mao2021one}. Specifically, we feed 60k points into the \nickname{} similar to the setting on the Waymo Dataset, and train 80 epochs for a fair comparison. As shown in Table~\ref{tab:once_val}, our method yields the competitive performance among all baselines. This again verifies the superiority of the proposed component  and the efficiency of our method applied on the large-scale complicated LiDAR scenarios.}

\begin{table}[t]
    \centering
    \scalebox{0.76}{
    \begin{tabular}{r|r|c|c|c|c}
    \Xhline{2.0\arrayrulewidth}
        Method          & \tabincell{c}{\textit{Mem.}}   & \tabincell{c}{\textit{Paral.}} &\tabincell{c}{\textit{Speed$^{\bot}$}}  & \tabincell{c}{\textit{Speed$^{\top}$}} & \tabincell{c}{\textit{Input Scale}}   \\
    \Xhline{2.0\arrayrulewidth}
        PointPillars~\cite{lang2019pointpillars}    & 354 MB  & 28   & 48    & 58     & 2$ \sim $9k         \\
        SECOND~\cite{yan2018second}                 & 710 MB  & 14   & 30    & 40     & 11$ \sim $17k       \\
        TANet~\cite{liu2020tanet}                   & 3000 MB & 3    & 20    & 28     & $<$12k             \\
        3DSSD~\cite{yang20203dssd}                  & 502 MB  & 19   & 11    & 28     & 16384     \\
        PointRCNN~\cite{shi2019pointrcnn}           & 560 MB  & 18   & 10    & 14     & 16384               \\
        Part-$A^2$~\cite{shi2020points}                 & 702 MB  & 13   & 12    & 19     & 11$ \sim $17k     \\
        PV-RCNN~\cite{shi2020pv}                     & 1223 MB & 8    & 8     & 10     & 11$ \sim $17k     \\
        IA-SSD (Ours)                                     & \textbf{102 MB}  & \textbf{100}  & \textbf{23}/\textbf{48$^{\dagger}$}    & \textbf{83}     & 16384     \\
    \Xhline{2.0\arrayrulewidth}
    \end{tabular}
    }
    \vspace{-0.2cm}
    \caption{\qy{ Efficiency comparison of different methods on the KITTI \textit{validation} set. Here, ``Mem.'' and ``Paral.'' denote the GPU memory footprint per frame during inference and the maximum number of batches that can be parallelized on one RTX2080Ti~(11GB). ''Speed$^{\bot}$'', ''Speed$^{\top}$'' is inference speed when processing one frame or full-loaded GPU memory, $^{\dagger}$ means dividing the scene into four parallel parts to speed up the first sampling layer. For a fair comparison, we also report each input scale of voxels/points according to their official setting. }}
    \label{tab:effi_kitti}
    \vspace{-0.5cm}
\end{table}

\qy{\textbf{Efficiency of \nickname{}.} Next, we evaluate the computational and memory efficiency of the proposed \nickname{}. In light of the performance variations on different hardware configurations, we re-implemented several representative approaches and report the memory and speed on the same platform for a fair comparison. Note that, we report the memory consumption by feeding the same input point cloud with 16384 points following the OpenPCDet configuration. For speed evaluation, different models are inference with batch point clouds with the full utilization of the same hardware. \revise{As shown in Table \ref{tab:effi_kitti}, the proposed \nickname{} has the lowest GPU memory consumption (up to 100 frames in parallel) with the highest inference speed (83 FPS) compared with existing benchmark approaches. }}
% To adapt to the steady improving of LiDAR sensor hardware, 
% \yf{Extensive experiments on Waymo and ONCE are also evaluated. As shown in Table~\ref{tab:effi_waymo&once}, our \nickname{} can also keep high efficiency with large-scale points input. This further demonstrates that our \nickname{} is highly efficient and suitable for real-time applications in vast autonomous driving panoramas.}}}
%\yf{Extensive experiments on Waymo~\cite{sun2020scalability} and ONCE~\cite{mao2021one} are also evaluated in Table~\ref{tab:effi_waymo&once}, which further demonstrates that our \nickname{} is highly efficient and suitable for real-time applications in vast autonomous driving panoramas.}}}

\subsection{Ablation Experiments}
\qy{In the following ablation studies, we train our \nickname{} with multi-class objects in a single model, and all experiments are conducted on the KITTI validation set.}

\textbf{Ablation on Downsampling Strategies.} \qy{To further verify the effectiveness of the proposed instance-aware sampling, we replace it with the D-FPS and Feat-FPS. As shown in Table~\ref{tab:ab_sample}, the proposed instance-aware sampling achieves the best detection performance in all three categories, especially the small objects such as \textit{pedestrians} and \textit{cyclists}. This shows that the proposed sampling strategy can effectively preserve the foreground information during down-sampling process, thereby achieving better detection. We also find that centroid-aware sampling (row 5) performs better on \textit{pedestrians} and \textit{cars}, but slightly inferior in detect \textit{cyclist} compared with mixed sampling  (row 4), primarily because this sampling strategy focus on the instance center, hence tends to ignore the distal geometric details of objects such as \textit{cyclist} with large aspect ratios.
% \yfcam{Specifically, Ctr-aware sampling (row 5) focuses more on the instance center, hence in favor of the \textit{pedestrian} class which is usually on small spatial scales. However, for \textit{cyclist} class with large aspect ratios, ctr-aware sampling still focus on the instance center and tends to ignore the distal geometric details of the objects, hence achieving inferior performance compared with mixed sampling strategies (Row 4). }
}

\begin{table}[tb]
\centering
   \resizebox{0.48\textwidth}{!}{   
   % \begin{tabular}{cccccc|ccc|ccc|ccc|}
   \begin{tabular}{p{1mm}<{\centering} p{6mm}<{\centering} p{6mm}<{\centering} p{6mm}<{\centering} p{6mm}<{\centering} p{3mm}<{\centering} ||p{14mm}<{\centering} |p{14mm}<{\centering} |p{14mm}<{\centering}}
\Xhline{2.0\arrayrulewidth}
   \multirow{2}{*}{\rotatebox{18}{}}& \multirow{2}{*}{\rotatebox{18}{D-FPS}}& \multirow{2}{*}{\rotatebox{18}{Feat-FPS}}& \multirow{2}{*}{\rotatebox{18}{Cls-aware}}& \multirow{2}{*}{\rotatebox{18}{Ctr-aware}} &
   & \small{Car Mod} & \small{Ped. Mod} & \small{Cyc. Mod}  \\
   % & & & & & & & Easy & Mod. & Hard & Easy & Mod. & Hard & Easy & Mod. & Hard  \\
   & & & & & & (IoU=0.7) & (IoU=0.5) & (IoU=0.5)  \\
\Xhline{2.0\arrayrulewidth}
  (1) & \checkmark & - & - & - & & 78.12 & 50.46 & 65.19  \\
%   (2) & - & \checkmark & - & & 78.88 & 10.45 & 16.91  \\
%   (3) & - & \checkmark & - & & 78.82 & 53.04 & 69.13  \\
  (2) & \checkmark & \checkmark & - & - & & 79.00 & 54.31 & 71.08  \\
%   (3) & - & - & \checkmark & & 79.27 & 24.90 & 37.12  \\
%   \left(6\right) & - & - & \checkmark & & 79.26 & 58.04 & \underline{71.20}  \\
%   \left(7\right) & - & - & \checkmark & & \underline{79.37} & \textbf{59.36} & 68.16  \\
  (3) & \checkmark & - & \checkmark & - & & 79.19 & 58.81 & 70.15  \\
  (4) & \checkmark & - & \checkmark & \checkmark & & 79.54 & 58.49 & \underline{71.33}  \\
  (5) & \checkmark & - & - & \checkmark & & {\underline{\textbf{79.57}}} & \underline{\textbf{58.91}} & \textbf{71.24}  \\
   \Xhline{2.0\arrayrulewidth}
   \end{tabular}} 
   \vspace{-0.2cm}
   \caption{\qy{Ablation study of \nickname{} on different sampling strategies, in which we report the 3D mAP with 11 recalls. Here D-FPS is the traditional Farthest Point Sampling, Feat-FPS represents the Feature-based FPS and Cls/Ctr-aware denote the proposed two learning based sampling methods.}}
   \label{tab:ab_sample}
   \vspace{-0.4cm}
\end{table}

\textbf{Ablation on Contextual Centroid Perception.} \qy{We further validate the effectiveness of the proposed contextual centroid perception module. Here, by replacing this module with the vanilla center-assign\yfcam{\footnote{https://github.com/qiqihaer/3DSSD-pytorch-openPCDet}} %\yfcam{\footnote{\yfcam{Noted that the 3DSSD can get better performance on \textit{car} class when applying center-assign in our Pytorch version, consequently we use center-assign to refer to the 3DSSD without special instructions in this work.}}} 
or original-assign used in \cite{qi2019deep}, the detection performance shows clearly decrease. \revise{Table~\ref{tab:ab_vote} shows that the inclusion of contextual points can indeed improve the detection performance, especially for small objects, since the representative points lie in the ground-truth bounding boxes are actually quite limited}. We are also noticed that both the extend-factor (2$\times$ size for each bounding box) and extend-length ($+$1.0m for each bounding box) have their own advantage in specific categories, showing that different contextual information may have a varying impact on different objects.}

\begin{table}[tb]
\centering
   \resizebox{0.48\textwidth}{!}{   
   % \begin{tabular}{cccccc|ccc|ccc|ccc|}
%   \begin{tabular}{p{2mm}<{\centering} p{5mm}<{\centering} p{5mm}<{\centering} p{5mm}<{\centering} p{5mm}<{\centering} p{8mm}<{\centering} ||p{14mm}<{\centering} |p{14mm}<{\centering} |p{14mm}<{\centering}}
% \Xhline{2.0\arrayrulewidth}
%   \multirow{2}{*}{\rotatebox{18}{}}
%   & \multirow{2}{*}{\rotatebox{18}{Centers-assign}} & \multirow{2}{*}{\rotatebox{18}{Origin-assign}} & \multirow{2}{*}{\rotatebox{18}{Extend-factor}} & \multirow{2}{*}{\rotatebox{18}{Extend-length}} & 
%   & \small{Car Mod} & \small{Ped. Mod} & \small{Cyc. Mod} \\
%   % & & & & & & & Easy & Mod. & Hard & Easy & Mod. & Hard & Easy & Mod. & Hard  \\
%   & & & & & & (IoU=0.7) & (IoU=0.5) & (IoU=0.5) \\
% \Xhline{2.0\arrayrulewidth}
% %   (1) & - & - & \checkmark & - &  & 78.12 & 50.46 & 65.19  \\
% %   (2) & \checkmark & - & - & - &  & 78.88 & 10.45 & 16.91  \\
% %   (3) & - & \checkmark & - & - &  & 78.82 & 53.04 & 69.13  \\
% %   (4) & - & - & \checkmark & - &  & 79.00 & 54.31 & 71.08  \\
%   (1) & \checkmark & - & - & - &  & 79.27 & 24.90 & 37.12  \\
% %   (2) & - & \checkmark & - & - &  & 79.26 & 58.04 & 71.20  \\
%   (2) & - & \checkmark & - & - &  & \todo{79.18} & \todo{55.62} & \todo{69.37}  \\
%   (3) & - & - & \checkmark & - &  & 79.37 & 58.36 & 68.16  \\
%   (4) & - & - & - & \checkmark &  & {\underline{\textbf{79.57}}} & \underline{\textbf{58.91}} &
  
    \begin{tabular}{p{2mm}<{\centering} p{43mm}<{\centering} ||p{14mm}<{\centering} |p{14mm}<{\centering} |p{14mm}<{\centering}}
\Xhline{2.0\arrayrulewidth}
   
   & \multirow{2}{*}{Centroid Perception Type} & \small{Car Mod} & \small{Ped. Mod} & \small{Cyc. Mod} \\
   % & & & & & & & Easy & Mod. & Hard & Easy & Mod. & Hard & Easy & Mod. & Hard  \\
   & & (IoU=0.7) & (IoU=0.5) & (IoU=0.5) \\
\Xhline{2.0\arrayrulewidth}
%   (1) & - & - & \checkmark & - &  & 78.12 & 50.46 & 65.19  \\
%   (2) & \checkmark & - & - & - &  & 78.88 & 10.45 & 16.91  \\
%   (3) & - & \checkmark & - & - &  & 78.82 & 53.04 & 69.13  \\
%   (4) & - & - & \checkmark & - &  & 79.00 & 54.31 & 71.08  \\
  (1) & Centers-assign &  79.27 & 24.90 & 37.12  \\
%   (2) & - & \checkmark & - & - &  & 79.26 & 58.04 & 71.20  \\
  (2) & Origin-assign  &  \todo{79.18} & \todo{55.62} & \todo{69.37}  \\
  (3) & Extend-factor assign &  79.37 & 58.36 & 68.16  \\
  (4) & Extend-length assign &  {\underline{\textbf{79.57}}} & \underline{\textbf{58.91}} & \underline{\textbf{71.24}}  \\
   \Xhline{2.0\arrayrulewidth}
   \end{tabular}} 
   \vspace{-0.2cm}
   \caption{Ablation study of \nickname{} framework with different centroid perception strategies.}
   \label{tab:ab_vote}
   \vspace{-0.4cm}
\end{table}

\section{Conclusion}
\label{sec:conclusion}
% \qy{In this paper, we propose an efficient single-stage framework called \nickname{} for 3D object detection. Considering the task of object detection inherently focuses on the foreground points, we propose an instance-aware learning-based downsampling way to automatically select the sparse yet important instance points. In addition, a dedicated contextual centroid perception module is proposed to fully exploit the geometrical structure around the bounding boxes. Extensive experiments conducted on the KITTI detection benchmark demonstrated the superior efficiency and accuracy of the proposed \nickname{}. \revise{In future work, we will further tackle extreme cases such as overlapped bounding boxes.}}

%This paper presents a new point-based single-stage 3D object detection networks, named \nickname{}. With novel instance-aware downsampling strategy and centroid rally module, we can effectively and efficiently achieve muti-class 3D object detection in a bottom-up manner.  Our \nickname{} achieves the best results among pure point-based methods, and provides a state-of-the-art efficiency than existing LiDAR detectors. In the future, we will focus on designing an efficient network to achieve real-time and robust 3D detection in 360-degree LiDAR scenes.

\qy{In this paper, we propose an efficient solution termed \nickname{} for point-based 3D object detection in LiDAR point clouds. Considering the task of object detection inherently focuses on the foreground information, we propose an instance-aware learning-based downsampling way to automatically select the sparse yet important instance points. Additionally, a dedicated contextual centroid perception module is proposed to fully exploit the geometrical structure around the bounding boxes. Extensive experiments conducted on three detection benchmarks demonstrated the superior efficiency and accuracy of the proposed \nickname{}. 
}

\smallskip\noindent\textbf{Limitations.} Although the proposed \nickname{} can achieve remarkable efficiency in object detection of large-scale LiDAR points clouds, it also has limitations. \textit{e.g.,} the instance-aware sampling relies on the semantic prediction of each point, which is susceptible to class imbalances distribution. For future work, we will further explore advanced techniques to alleviate the imbalanced issue.

\smallskip\noindent\textbf{Acknowledgements.} This work was partially supported by the National Natural Science Foundation of China (No. 61972435, U20A20185), China Scholarship Council (CSC) scholarship, and Huawei UK AI Fellowship.

\clearpage
%%%%%%%%% REFERENCES
{\small
\bibliographystyle{ieee_fullname}
\bibliography{egbib}
}

\newpage
\clearpage
\appendix
\section*{Appendix}

\section{Details of The Proposed \nickname{}}

\smallskip\noindent\textbf{(1) Detailed Network Architecture.}
% \qy{We provide the detailed architecture of our \nickname{} here. The network follows the encoder-only framework 3DSSD \cite{yang20203dssd} but has a more lightweight backbone setting. Details of the architecture on KITTI are as follows:
% }
%\yf{We provide the detailed architecture of our \nickname{} here. The backbone consists of three SA layers~\cite{qi2017pointnet++} with only two radius for query. Details of the architecture deployed on KITTI Dataset are as follows:}
\qy{Here, we provide the detailed architecture of our \nickname{}. The proposed \nickname{} has a lightweight backbone, which consists of three SA (Set Abstraction) layers~\cite{qi2017pointnet++} with only two radii for the spherical neighbor query. The detailed architecture deployed on KITTI Dataset is as follows: }
\vspace{0.2cm}

\noindent syntax: $SA(npoint, [radii], [nquery], [dimension])$
\yf{
\vspace{-0.05cm}
\begin{equation*}
\begin{aligned}
&SA(4096, [0.2,0.8], [16,32], [[16,16,32],[32,32,64]])  \\
&\rightarrow MLP(96 \rightarrow 64) \\
&SA(1024, [0.8,1.6], [16,32], [[64,64,128],[64,96,128]]) \\
&\rightarrow MLP(256 \rightarrow 128) \\
&SA(512, [1.6,4.8], [16,32], [[128,128,256],[128,256,256]]) \\
&\rightarrow MLP(512 \rightarrow 256) \\
\end{aligned}
\end{equation*}
}

\noindent where $npoint$ denotes the number of sampled points, $[radii]$ denote the grouping radii, $[nquery]$ denotes the number of grouping points, $[dimension]$ denotes the feature dimensions.

\qy{The class/centroid-aware prediction layer:}
\begin{equation*}
\begin{aligned}
&MLP(256 \rightarrow 256 \rightarrow 3) \\
\end{aligned}
\end{equation*}

The architecture of the contextual instance centroid perception module is as follows:
\begin{equation*}
\begin{aligned}
&MLP(256 \rightarrow 128 \rightarrow 3) \\
\end{aligned}
\end{equation*}

The architecture of centroid-based instance aggregation is as follows:
\yf{
\begin{equation*}
\begin{aligned}
&SA(256, [4.8,6.4], [16,32], [[356,356,512],[256,512,1024]]) \\
&\rightarrow MLP(1536 \rightarrow 512) \\
\end{aligned}
\end{equation*}
}
The final detection head is composed of two branches:
\begin{equation*}
\begin{aligned}
&\textit{cls~branch}: FC(512) \rightarrow FC(256) \rightarrow FC(256) \rightarrow FC(3)  \\
&\textit{reg~branch}: FC(512) \rightarrow FC(256) \rightarrow FC(256) \rightarrow FC(30) \\
\end{aligned}
\end{equation*}

\qy{Considering the large-scale spatial ranges and increasing number of potential instances in the Waymo and ONCE datasets, the number of sampled points are improved to 16384, 4096, 2048, and 1024 in our framework, and the contextual centroid perception boundary is improved to 2.0m. The rest of the hyperparameters are kept consistent for a fair comparison.}

\section{Additional Implementation Details}

\smallskip\noindent\textbf{(1) Data augmentation.} \qy{During training, We also apply two data augmentation strategies including scene-level augmentation and object-level augmentation. The detailed settings and hyperparameters are as follows:}

\noindent{\textbf{Scene-level augmentation:}}
% \vspace{-0.1cm}
\begin{itemize}
\setlength{\parsep}{0pt} 
\setlength{\topsep}{0pt} 
\setlength{\itemsep}{0pt}
\setlength{\parsep}{0pt}
\setlength{\parskip}{0pt}
    \item Random scene flip with a 50 $\%$ probability.
    \item Random scene rotation around $z$-axis with a random value from $[-\frac{\pi}{4}, \frac{\pi}{4}]$.
    \item Random scene scaling with a random factor from $[0.95, 1.05]$.
    % \item Transform objects (car: 20, pedestrian: 15, cyclist: 15) from other scenes, note that the internal points of all sampled objects should be over 5.
\end{itemize}

\noindent{\textbf{Object-level augmentation:}}
\vspace{-0.1cm}
\begin{itemize}
\setlength{\parsep}{0pt} 
\setlength{\topsep}{0pt} 
\setlength{\itemsep}{0pt}
\setlength{\parsep}{0pt}
\setlength{\parskip}{0pt}
    \item \qy{Transform objects from other scenes. In particular, 20 cars, 15 pedestrians, and 15 cyclists are copied to the current scene. Note that, the minimum number of points for a sampled instance is 5.}
\end{itemize}

\noindent \textbf{(2) Training and inference.}  \qy{We train the proposed \nickname{} in an end-to-end fashion with a maximum of 80 epochs. Adam solver with onecycle learning strategy \cite{smith2019super} is used for optimization. In our experiment, the batch size is set to 8, and the learning rate is set to 0.01. During inference, our \nickname{} is able to take raw point clouds and generate proposals for all objects in a single forward pass. Finally, all proposals are filtered by 3D-NMS post-processing with an IoU threshold of 0.01 on KITTI and 0.1 on Waymo/ONCE.}

% \qy{The proposed \nickname{} is built upon OpenPCDet \cite{openpcdet2020} with default settings. It trained end-to-end for a maximum epoch number of 80. We use Adam optimizer with onecycle learning strategy \cite{smith2019super}. In our experiment, the batch size is set to 8, and the learning rate is set to 0.01. During inference, our \nickname{} is able to take raw point clouds and generate proposals for all objects in a single forward pass. Finally, all proposals are filtered by 3D-NMS post-processing with an IoU threshold of 0.1.}
%\yf{We train the proposed \nickname{} end-to-end for 80 epochs with the Adam optimizer with onecycle learning strategy \cite{smith2019super}. In our experiment, the batch size is set to 8, and the learning rate is set to 0.01. During inference, our \nickname{} is able to take raw point clouds and generate proposals for all objects in a single forward pass. Finally, all proposals are filtered by 3D-NMS post-processing with an IoU threshold of 0.1 on KITTI/Waymo and 0.01 on ONCE.}

\begin{table*}[t]
  \begin{center}
      \resizebox{0.93\textwidth}{!}{   
      % \begin{tabular}{cccccc|ccc|ccc|ccc|}
      \begin{tabular}{p{16mm}<{\centering} p{16mm}<{\centering} p{16mm}<{\centering} p{16mm}<{\centering} ||p{14mm}<{\centering} |p{14mm}<{\centering} |p{14mm}<{\centering} ||p{16mm}<{\centering} |p{16mm}<{\centering} |p{16mm}<{\centering}}
\Xhline{2.0\arrayrulewidth}
      \multirow{2}{*}{$1^{st}$ layer}& \multirow{2}{*}{$2^{nd}$ layer}& \multirow{2}{*}{$3^{rd}$ layer}& \multirow{2}{*}{$4^{th}$ layer} & \multirow{2}{*}{Recall Car} & \multirow{2}{*}{Recall Ped.} & \multirow{2}{*}{Recall Cyc.} & \small{Car Mod} & \small{Ped. Mod} & \small{Cyc. Mod}  \\
      & & & & & & & (IoU=0.7) & (IoU=0.5) & (IoU=0.5)  \\
\Xhline{2.0\arrayrulewidth}
      Random & Random & Random & Random & 67.4$\%$ & 72.1$\%$ & 57.3$\%$ & 75.02  & 51.16 & 66.07 \\
      D-FPS & D-FPS & D-FPS & D-FPS & 91.4$\%$ & 69.1$\%$ & 71.6$\%$ & 78.12 & 50.46 & 65.19  \\
      D-FPS &  Feat-FPS &  Feat-FPS & Feat-FPS & 95.3$\%$ & 80.1$\%$ & 91.7$\%$ & 79.00 & 54.31 & 71.08 \\
    %   D-FPS & Ctr-aware & Ctr-aware & Ctr-aware & 75.5$\%$ & 79.2$\%$ & 78.9$\%$ & 68.05 & 54.42 & 68.67  \\
      D-FPS & D-FPS & Cls-aware & Cls-aware & 97.9$\%$ & 97.4$\%$  & 92.7$\%$ & 79.19 & 58.81 & 70.15  \\
      
      D-FPS & D-FPS & Cls-aware & Ctr-aware & 97.9$\%$ & 97.7$\%$ & 96.3$\%$ & 79.54 & 58.49 & \underline{71.33} \\
      \textbf{D-FPS} & \textbf{D-FPS} & \textbf{Ctr-aware} & \textbf{Ctr-aware} & \textbf{97.9$\%$} & \underline{\textbf{98.4$\%$}} & \textbf{97.2$\%$} & \underline{\textbf{79.57}} & \underline{\textbf{58.91}} & \textbf{71.24} \\
\Xhline{2.0\arrayrulewidth}
  \end{tabular}}
  \end{center}
  \vspace{-0.4cm}
  \caption{\revise{The correlation between the instance recall ratio and the final detection performance.}}
  \label{tab:ins_recall_detail}
  \vspace{-0.2cm}
  \end{table*}

\section{Additional Experimental Results}

\begin{table}[h]
   \centering
   \begin{minipage}[t]{0.49\textwidth}
   \begin{center}
   \resizebox{0.9\textwidth}{!}{ 
  \begin{tabular}{p{2.2cm}<{\centering}| p{1.3cm}<{\centering}| p{1.3cm}<{\centering} |p{1.3cm}<{\centering} |p{1.4cm}<{\centering} }
    \Xhline{2.0\arrayrulewidth}
   Method & 256p  & 1024p  & 4096p & 16384p \\
    \Xhline{2.0\arrayrulewidth}
    D-FPS \cite{qi2019deep} & \textless 0.1 ms & 0.5 ms & 2.8 ms & 23.7 ms \\
    Feat-FPS \cite{yang20203dssd} & 0.3 ms & 0.7 ms & 4.2 ms & 40.6 ms \\
    \textbf{Cls/Ctr-aware} & \textbf{0.2 ms} & \textbf{0.2 ms} & \textbf{0.3 ms} & \textbf{0.5 ms}  \\
%   Memory Consumption (MB) & 478 & 980 & 544  \\
%   Speed (\todo{fps}) & 56  & 40 & 48  \\
    \Xhline{2.0\arrayrulewidth}
   \end{tabular}}
   \end{center}
   \vspace{0.1cm}
%   \caption{The computational memory consumption of different sampling methods.}
   \end{minipage}
   
\vspace{-0.2cm}
   \begin{minipage}[t]{0.49\textwidth}
   \begin{center}
   \resizebox{0.9\textwidth}{!}{ 
%   \begin{tabular}{c|c|c|c|c|c}
  \begin{tabular}{p{2.2cm}<{\centering}| p{1.3cm}<{\centering}| p{1.3cm}<{\centering} |p{1.3cm}<{\centering} |p{1.4cm}<{\centering} }
    \Xhline{2.0\arrayrulewidth}
   Method & 256p & 1024p & 4096p & 16384p \\
    \Xhline{2.0\arrayrulewidth}
    D-FPS \cite{qi2019deep} & \textless 1 MB & \textless 1 MB & \textless 1 MB & \textless 1 MB \\
    Feat-FPS \cite{yang20203dssd} & 64 MB & 104 MB & 448 MB & 6228 MB \\
    \textbf{Cls/Ctr-aware} & \textbf{0.25 MB} & \textbf{1 MB} & \textbf{4 MB} & \textbf{17 MB}  \\
%   Memory Consumption (MB) & 478 & 980 & 544  \\
%   Speed (\todo{fps}) & 56  & 40 & 48  \\
    \Xhline{2.0\arrayrulewidth}
   \end{tabular}}
   \end{center}
   \end{minipage}
%   \vspace{-0.2cm}
   \caption{\revise{Time and memory consumption of sampling methods.}}
   \label{tab:effi_sample}
  \vspace{-0.3cm}
   \end{table}

\smallskip\noindent\textbf{(1) Preserving more foreground points really benefits the final detection performance?}  As mentioned in section \ref{subsec:instance-aware downsampling strategy}, two instance-aware strategies are proposed to keep high instance recall while hierarchically downsampling the points. However, it remains unclear that whether the more foreground points really benefit the final detection performance. To this end, we further justify the motivation of our \nickname{} here. Specifically, we conduct several groups of experiments based on our framework with different sampling strategies. Note that, the network architecture and parameter settings are kept consistent. The quantitative detection results, accompanied with the instance recall ratio after the last downsampling layers by using different possible combinations of the sampling approaches are shown in Table \ref{tab:ins_recall_detail}.

\qy{From the results in Table \ref{tab:ins_recall_detail} we can see that: (1) the instance recall ratio is positively correlated with the final detection performance, especially for small objects with a limited number of points such as \textit{pedestrians} and \textit{cyclists}. (2) The detection performance of \textit{cars} is relatively robust to the variations of sampling strategies, primarily because that \textit{car} usually has a sufficient number of foreground points remaining after downsampling, hence relatively easy to be detected. (3) Adopting the proposed instance-aware sampling strategies at the early encoding layers may negatively affect the final detection performance, primarily because of the insufficient semantic information in the early latent point features. (4) Deploying the proposed instance-aware downsampling strategies at the last two encoding layers can significantly improve the detection performance. Overall, this experiment further demonstrates that more foreground points are appealing for object detection task, especially for small but important objects.
}

\smallskip\noindent\textbf{(2) Efficiency of Sampling.}  
We further explore the efficiency of different sampling strategies, to have an intuitive idea of the advantages of our instance-aware sampling. Table~\ref{tab:effi_sample} compares the time and memory consumption of different sampling strategies with a varying number of points. 
% Note that, we only change the sampling strategies while keeping all other modules unchanged. 
We can clearly see that the proposed instance-aware sampling has superior efficiency compared with the Feat-FPS \cite{yang20203dssd}, hence leading to a higher frame rate of our method during inference.
%\qy{From the results in Table \ref{tab:ins_recall_detail} we can also see that: {1) Adopting the proposed instance-aware sampling strategies at the early encoding layers may negatively affect the final detection performance, primarily because of the insufficient semantic information in the early latent point features.} 2) Deploying the proposed instance-aware downsampling strategies at the last two encoding layers can significantly improve the detection performance. Additionally, adopting the centroid-aware sampling at the final encoding layer is likely to achieve slightly better results, since the last feature map is quite sparse and informative. Preserving more centroid points is likely to ease the difficulty in the centroid perception module. }

%\qy{From the results in Table \ref{tab7} we can see that: /todo{1) Adopting the proposed instance-aware sampling strategies at the early encoding layers may negatively affect the final detection performance, primarily because of the insufficient semantic information in the early latent point features.} 2) Deploying the proposed instance-aware downsampling strategies at the last two encoding layers can significantly improve the detection performance. Additionally, adopting the centroid-aware sampling at the final encoding layer is likely to achieve slightly better results, since the last feature map is quite sparse and informative. Preserving more centroid points is likely to ease the difficulty in the centroid perception module. }

\smallskip\noindent\textbf{(3) Evaluation on KITTI validation set.} \qy{We also report the detection results achieved by several representative approaches on the \textit{validation} set of the KITTI Dataset in Table \ref{tab:kitti_val_pcdet}. Note that, all results achieved by baselines are reproduced based on the OpenPCDet\footnote{https://github.com/open-mmlab/OpenPCDet}. In particular, all baselines are trained with multi-class objects in a single model for a fair comparison. It can be seen that our single-stage \nickname{} achieves superior detection performance compared with other point-based baselines. We also noticed that the prior SoTA detector 3DSSD\footnote{https://github.com/qiqihaer/3DSSD-pytorch-openPCDet} achieve poor results on the class of pedestrian and cyclist, further demonstrating the advantages of our \nickname{}.}

\smallskip\noindent\textbf{(4) Efficiency of our \nickname{} on large-scale LiDAR scenarios.} To further verify the efficiency of our \nickname{} on large-scale 3D datasets, we further report the efficiency of our \nickname{} on the validation set of Waymo and ONCE datasets. As shown in Table~\ref{tab:effi_waymo&once}, the proposed \nickname{} can still achieve satisfactory real-time performance in such complex panoramic scenes.

%\smallskip\noindent\textbf{(3)Efficiency performance on challenging large-scale LiDAR scenarios.} To further certify the efficiency and adaptation on advance LiDAR sensors, we also illustrate efficiency details on Waymo and ONCE \textit{validation} set. As shown in Table~\ref{tab:effi_waymo&once}, our \nickname{} still have capacity of keeping light-weight and efficient even in complicated panoramic scenes.

\begin{table}[t]
   \begin{center}
      \resizebox{0.48\textwidth}{!}{   
      \begin{tabular}{c|r|c||p{14mm}<{\centering}|p{14mm}<{\centering}|p{14mm}<{\centering}}
    %   \begin{tabular}{p{4mm}<{\centering} |p{26mm}<{\centering} |p{20mm}<{\centering} |p{18mm}<{\centering} |p{18mm}<{\centering} |p{18mm}<{\centering}}
\Xhline{2.0\arrayrulewidth}
      \multirow{2}{*}{} & \multirow{2}{*}{Method} & \multirow{2}{*}{Type} & \small{Car Mod} & \small{Ped. Mod} & \small{Cyc. Mod}  \\
      % & & & & & & & Easy & Mod. & Hard & Easy & Mod. & Hard & Easy & Mod. & Hard  \\
      & & &  (IoU=0.7) & (IoU=0.5) & (IoU=0.5)  \\
\Xhline{2.0\arrayrulewidth}
    %   \multirow{3}{*}{\rotatebox{90}{Voxel-based}}
      \multirow{3}{*}{Voxel-based}
      & SECOND \cite{yan2018second}   & 1-stage  & 78.62 & 52.98 & 67.15  \\
      & PointPillars \cite{lang2019pointpillars}   & 1-stage  & 77.28  & 52.29 & 62.28  \\
      & Part-$A^2$  \cite{shi2020points}    & 2-stage  & 79.40  & \underline{60.05} & 69.90  \\
\Xhline{1.0\arrayrulewidth}
      \multirow{1}{*}{Point-Voxel}
      & PV-RCNN  \cite{shi2020pv} & 2-stage & \underline{83.61}  & 57.90 & 70.47  \\
\Xhline{1.0\arrayrulewidth}  
      \multirow{3}{*}{Point-based}
      & PointRCNN \cite{shi2019pointrcnn} & 2-stage & 78.70 & 54.41 & \underline{72.11}  \\
      & 3DSSD \cite{yang20203dssd}    & 1-stage  & 79.06 & 10.49 & 16.93  \\ % 
% \Xhline{2.0\arrayrulewidth}
\cline{2-6}
% \Xhline{1.0\arrayrulewidth}
      & \textbf{\nickname{} (Ours)} & \textbf{1-stage} & \textbf{79.57} & \textbf{58.91} & \textbf{71.24}  \\
\Xhline{2.0\arrayrulewidth}
   \end{tabular}}
   \end{center}
   \vspace{-0.2cm}
   \caption{Performance comparison of different detectors based on the OpenPCDet library. Note that, all detectors are trained with multi-class objects together, and the results are achieved by using a single detection model.}
   \label{tab:kitti_val_pcdet}
%   \vspace{-0.3cm}
   \end{table} 
   
\begin{table}[t]
    \centering
    \scalebox{0.8}{
    \begin{tabular}{r|r|r|r|r|r}
    \Xhline{2.0\arrayrulewidth}
        Dataset   & \tabincell{c}{\textit{Mem.}}   & \tabincell{c}{\textit{Paral.}} &\tabincell{c}{\textit{Speed$^{\bot}$}}  & \tabincell{c}{\textit{Speed$^{\top}$}} & \tabincell{c}{\textit{Input Scale}}   \\
    \Xhline{2.0\arrayrulewidth}
        Waymo~\cite{sun2020scalability}       & 626 MB    & 16    & 9$^{\dagger}$     & 14           & 81920    \\
                    & 433 MB    & 23    & 8$^{\dagger}$     & 20           & 65536    \\
    \Xhline{1.0\arrayrulewidth}
        ONCE~\cite{mao2021one}        & 401 MB    & 25    & 11$^{\dagger}$    & 21           & 60k    \\
    \Xhline{2.0\arrayrulewidth}
    \end{tabular}
    }
    % \vspace{-0.2cm}
    \caption{\qy{Efficiency of our \nickname{} on Waymo and ONCE Datasets. The number of input points to our framework is increased,  considering the large-scale panoramic scenes compared with KITTI. Here ``Mem.'' and ``Paral.'' denote the GPU memory footprint per frame during inference and the maximum number of batches that can be parallelized on one RTX2080Ti~(11GB). ''Speed$^{\bot}$'', ''Speed$^{\top}$'' is inference speed when processing one frame or full-loaded GPU memory. $^{\dagger}$ We divide the whole scene into four parallel parts in the first sampling layer.}}
    \label{tab:effi_waymo&once}
    \vspace{-0.3cm}
\end{table}

\begin{figure*}[thb]
   \centering
   \includegraphics[width=1.0\textwidth]{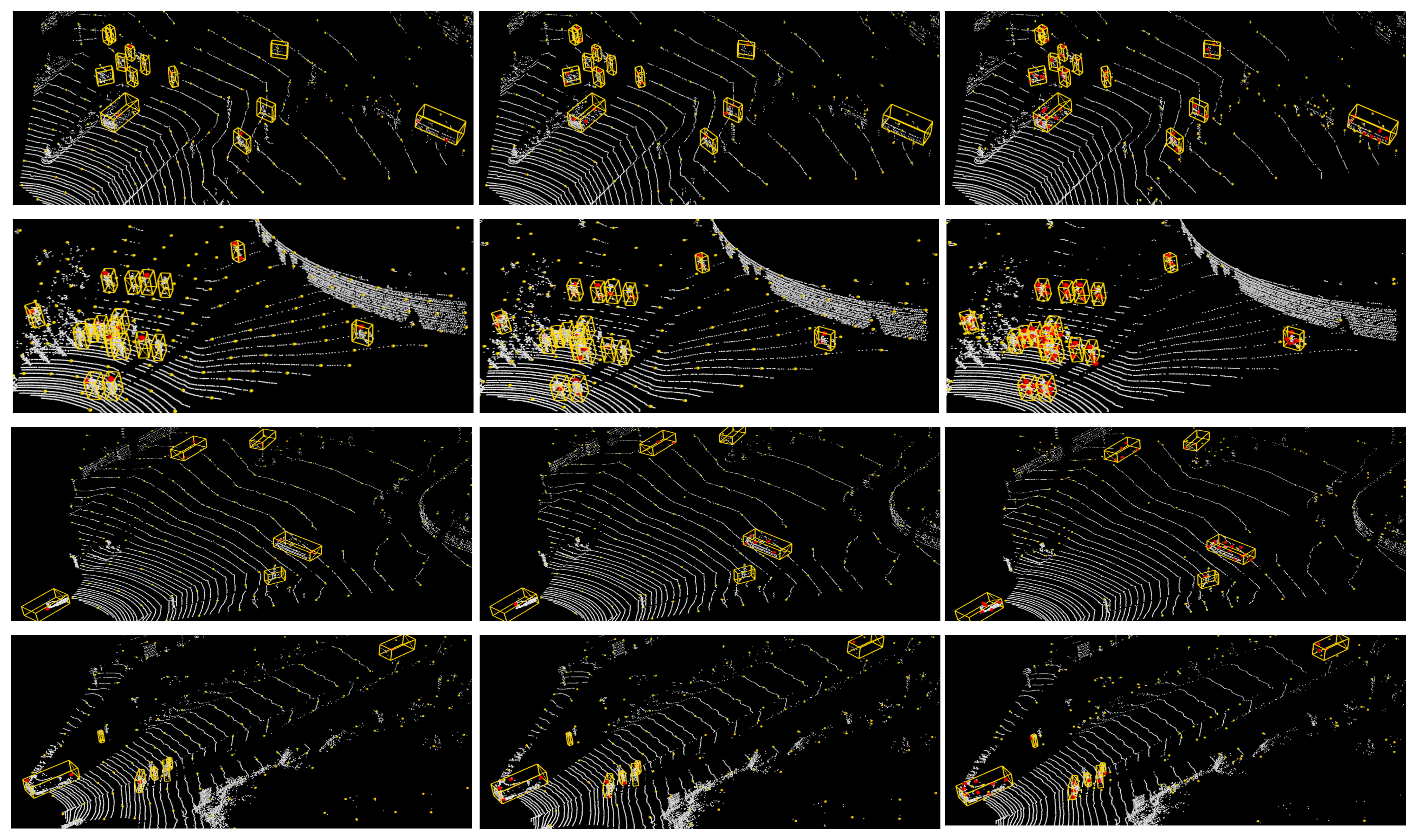}
    \caption{\qy{Qualitative visualization of the downsampled point clouds achieved by different sampling strategies (From left to right, D-FPS, F-FPS, and the proposed instance-aware sampling). Note that, the raw point clouds and representative points are colored in white and gold, respectively. Positive representative points are highlighted in red.}}
   \label{fig:visual_downdsample}
   \vspace{-0.5cm}
\end{figure*}

\smallskip\noindent\textbf{(5) Qualitative visualization of our instance-aware downsampling. }\qy{To intuitively compare the performance of different sampling approaches, we qualitatively show the visualization of the downsampled point clouds achieved by different approaches in Figure \ref{fig:visual_downdsample}. Clearly, the proposed instance-aware sampling can effectively preserve more foreground points (shown in red), especially for foreground points belonging to small and sparse instances (\textit{e.g.,} \textit{pedestrian}), as well instances far away from the sensors.}

\begin{figure*}[h]
   \centering
   \includegraphics[width=1.0\textwidth]{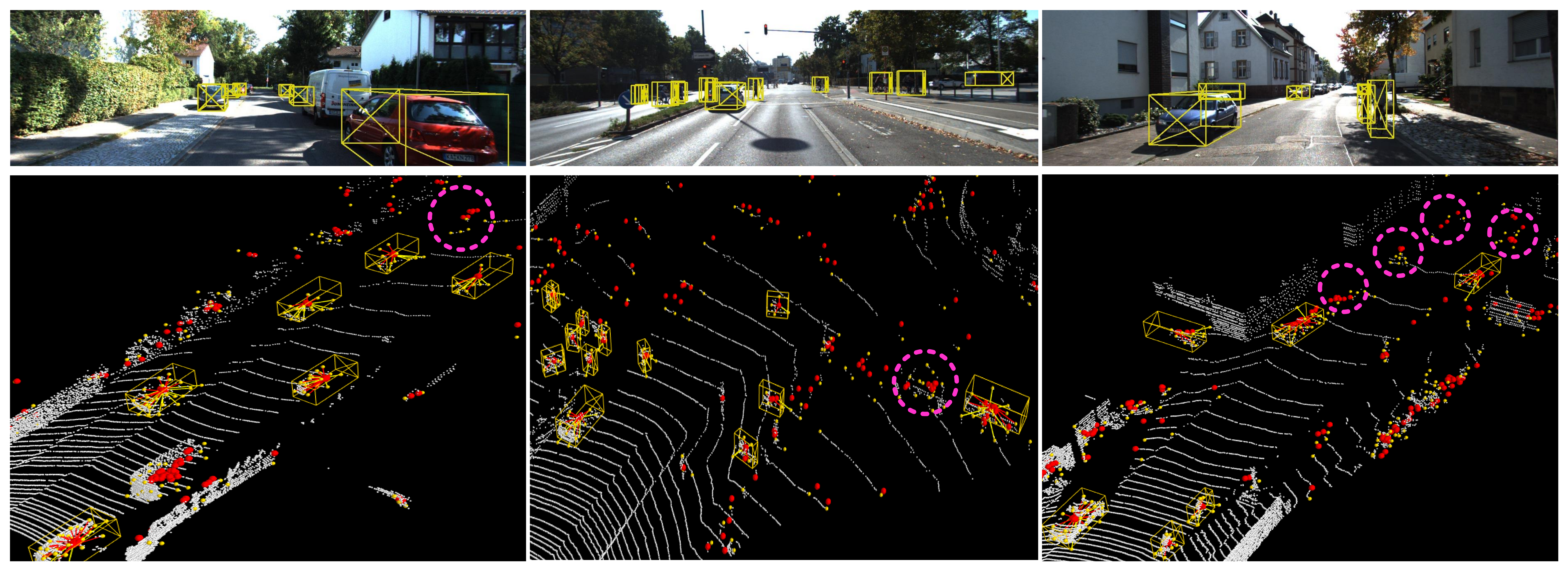}
    \vspace{-0.2cm}
    \caption{\qy{Visualization of the contextual centroid perception on the \textit{validation} spit of the KITTI dataset. All representative points and predicted centroid are colored in gold and red, respectively. In particular, we also show the offsets of representative points inside/around the objects in red/gold. Best viewed in color.}}
   \label{fig:visual_context_centroid}
   \vspace{-0.3cm}
\end{figure*}

\smallskip\noindent\textbf{(6) Visualization of the Contextual Centroid Perception.} \qy{We also visualize the results produced by our contextual centroid perception module in Figure \ref{fig:visual_context_centroid}. It is clear that the downsampled point clouds at this stage are quite sparse and insufficient, which makes the centroid estimation and instance regression considerably difficult. Therefore, it is necessary to exploit the useful information around the instance, even outside the ground-truth bounding boxes. Thanks to the proposed contextual centroid perception module, our \nickname{} can even precept the objects with extremely indistinguishable geometry and limited points (shown in purple dotted circles). This further demonstrated the effectiveness of the proposed module.}

\begin{figure*}[h]
   \centering
    \includegraphics[width=1.0\textwidth]{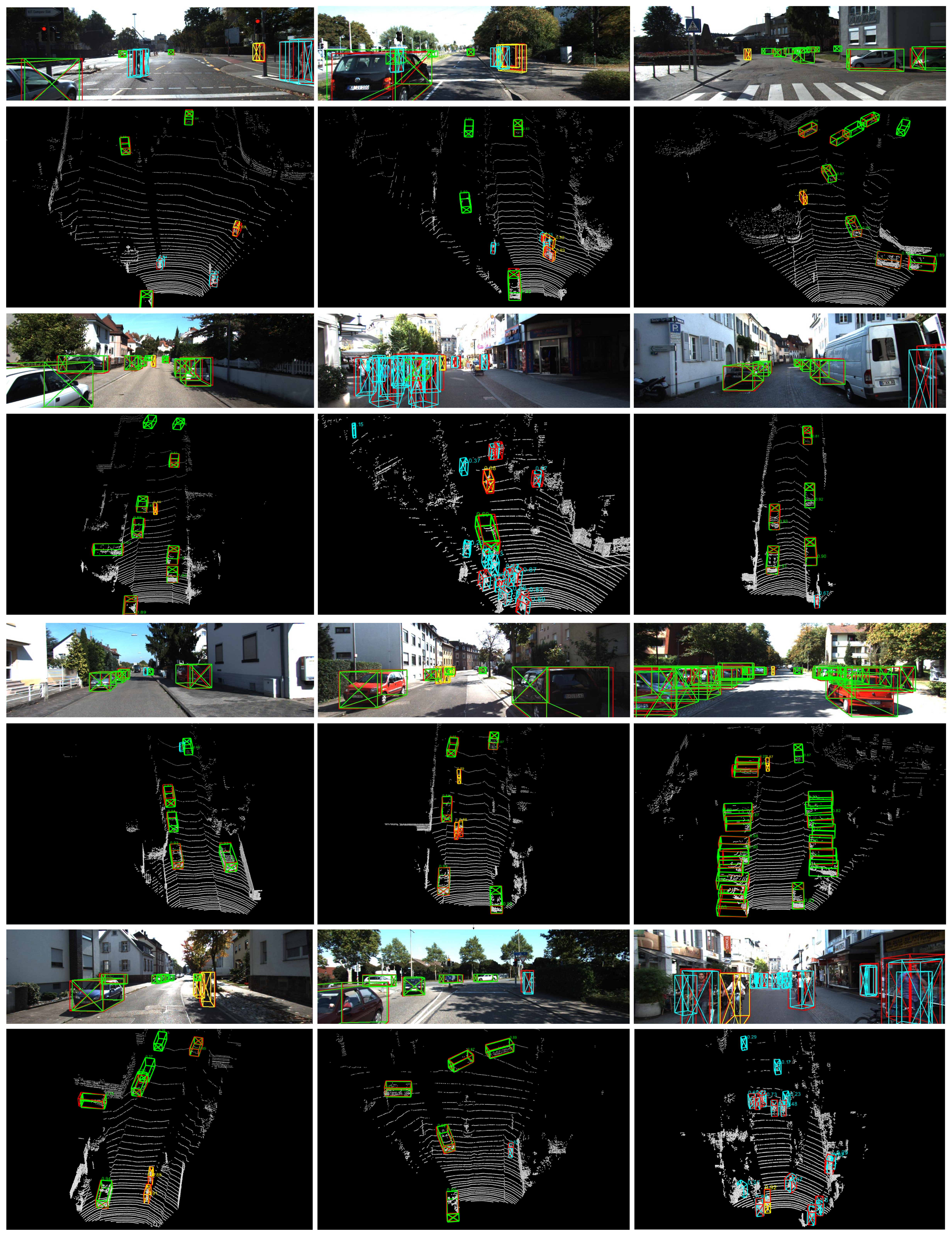}
    \caption{\qy{Extra qualitative results achieved by our \nickname{} on the \textit{validation} set of the KITTI Dataset. We also show the corresponding projected 3D bounding boxes on images. Note that, the ground-truth bounding boxes are shown in red, and the predicted bounding boxes are shown in green for \textit{car}, cyan for \textit{pedestrian}, and yellow for \textit{cyclist}. Best viewed in color.}}
   \label{fig:visual_kitti_val}
\end{figure*}

\begin{figure*}[h]
   \centering
   \includegraphics[width=1.0\textwidth]{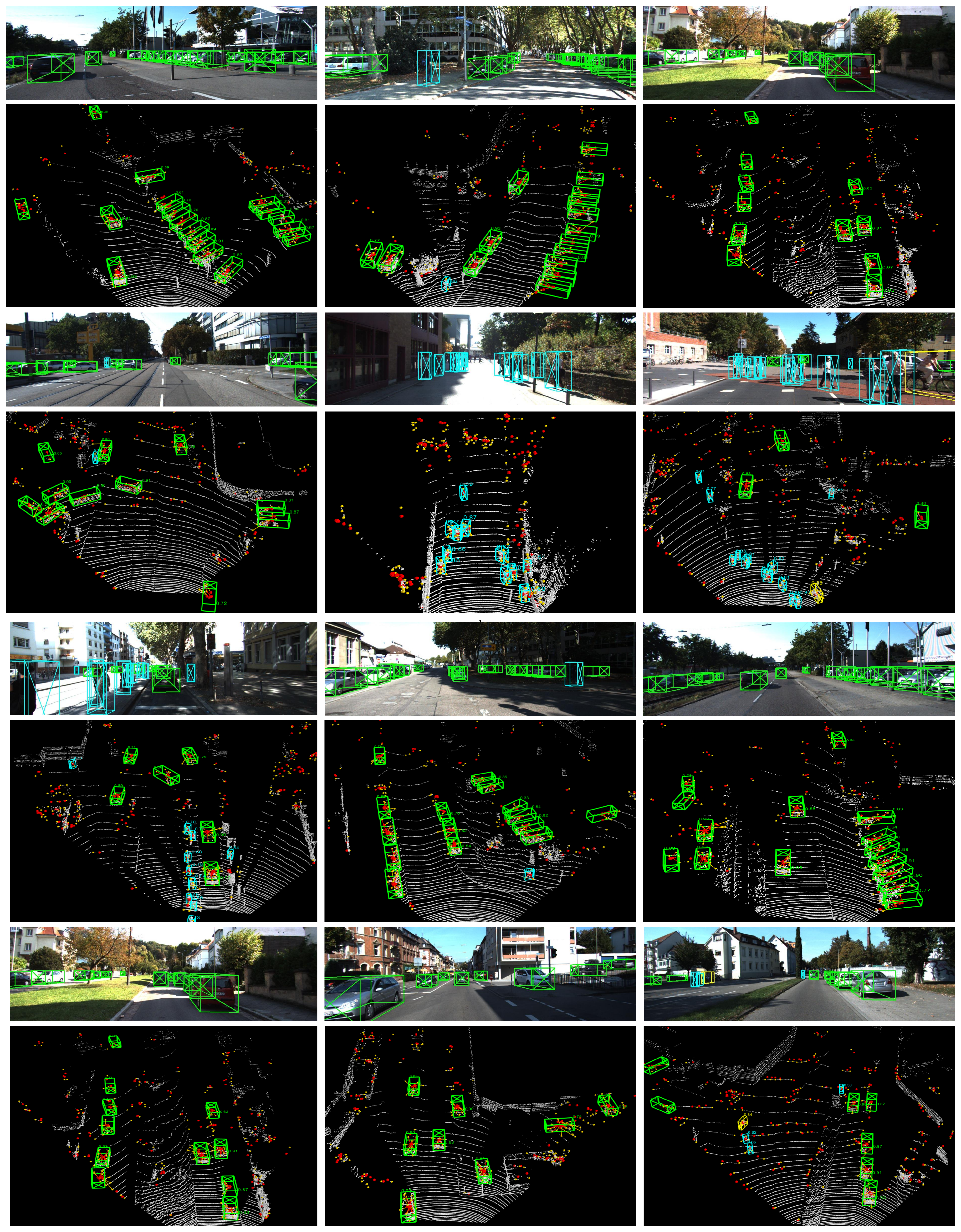}
    \caption{\qy{Extra qualitative results achieved by our \nickname{} on the \textit{test} set of the KITTI Dataset. We also show the corresponding projected 3D bounding boxes on images. Note that, there is no ground-truth bounding boxes available, hence we only show the predicted bounding boxes in green for \textit{car}, cyan for \textit{pedestrian}, and yellow for \textit{cyclist}. The centroid predictions are marked in red, while the 256 representative points are shown in gold. Best viewed in color.}}
   \label{fig:visual_kitti_test}
\end{figure*}

\smallskip\noindent\textbf{(7) Additional qualitative detection results on the KITTI Dataset.} \qy{We also show extra qualitative detection results achieved by our \nickname{} on the \textit{validation} (Figure \ref{fig:visual_kitti_val}) and \textit{test} (Figure \ref{fig:visual_kitti_test}) split of the KITTI Dataset. It can be seen that our \nickname{} can achieve satisfactory detection performance on this dataset, even for some challenging cases. It is also worth mentioning that the detection results of different objects are achieved by our \nickname{} in a single pass, instead of the common practice to train separate models for different objects.}

\begin{figure*}[thb]
   \centering
   \includegraphics[width=0.92\textwidth]{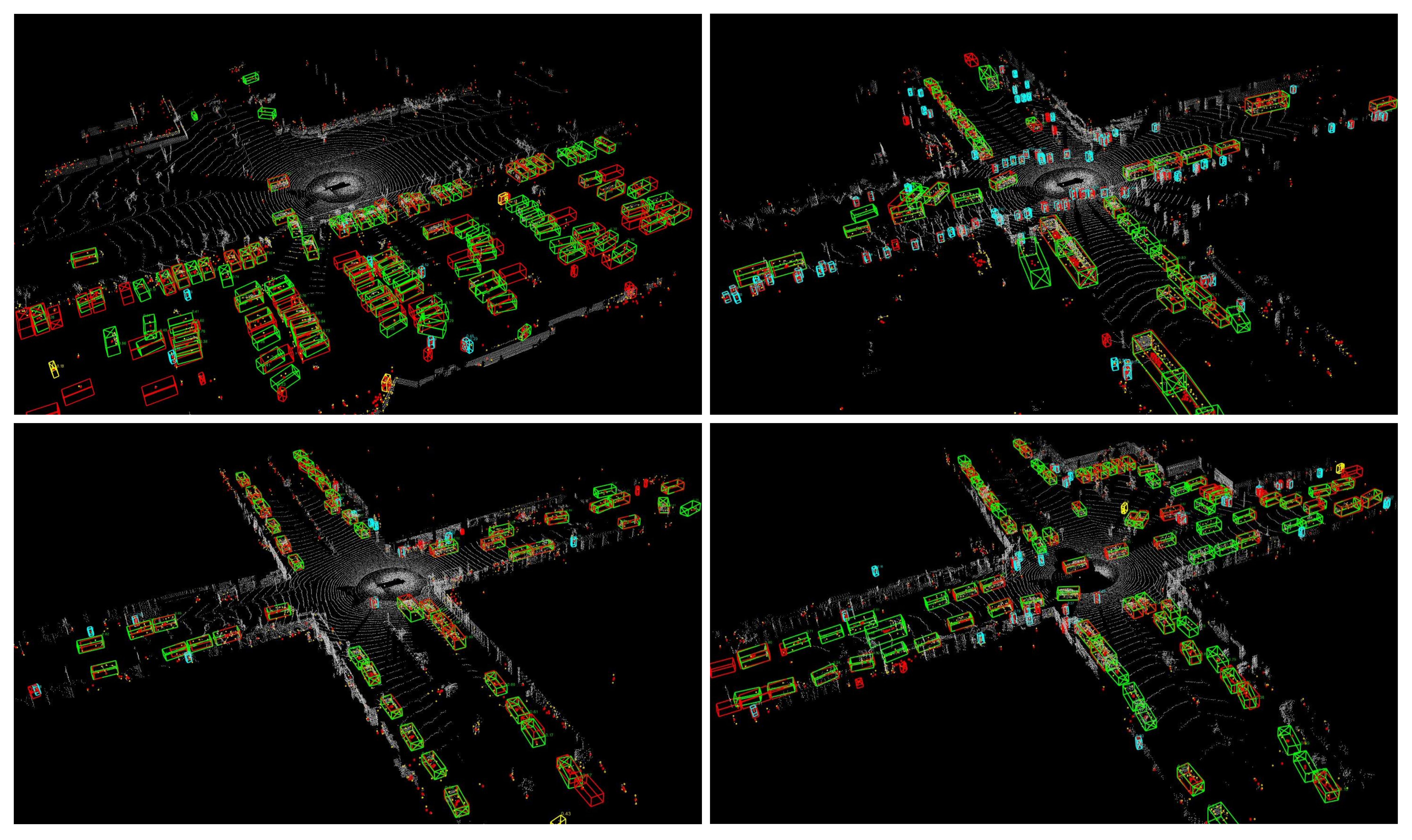}
    \caption{\yf{Extra qualitative results achieved by our \nickname{} on the \textit{val} set of the Waymo Dataset. Here We demonstrate our detection results on some challenging scenes. Note that, the ground-truth bounding boxes are shown in red, and the predicted bounding boxes are shown in green for \textit{vehicle}, cyan for \textit{pedestrian}, and yellow for \textit{cyclist}. Best viewed in color.}}
   \label{fig:visual_waymo_val}
\end{figure*}

\begin{figure*}[thb]
   \centering
   \includegraphics[width=0.92\textwidth]{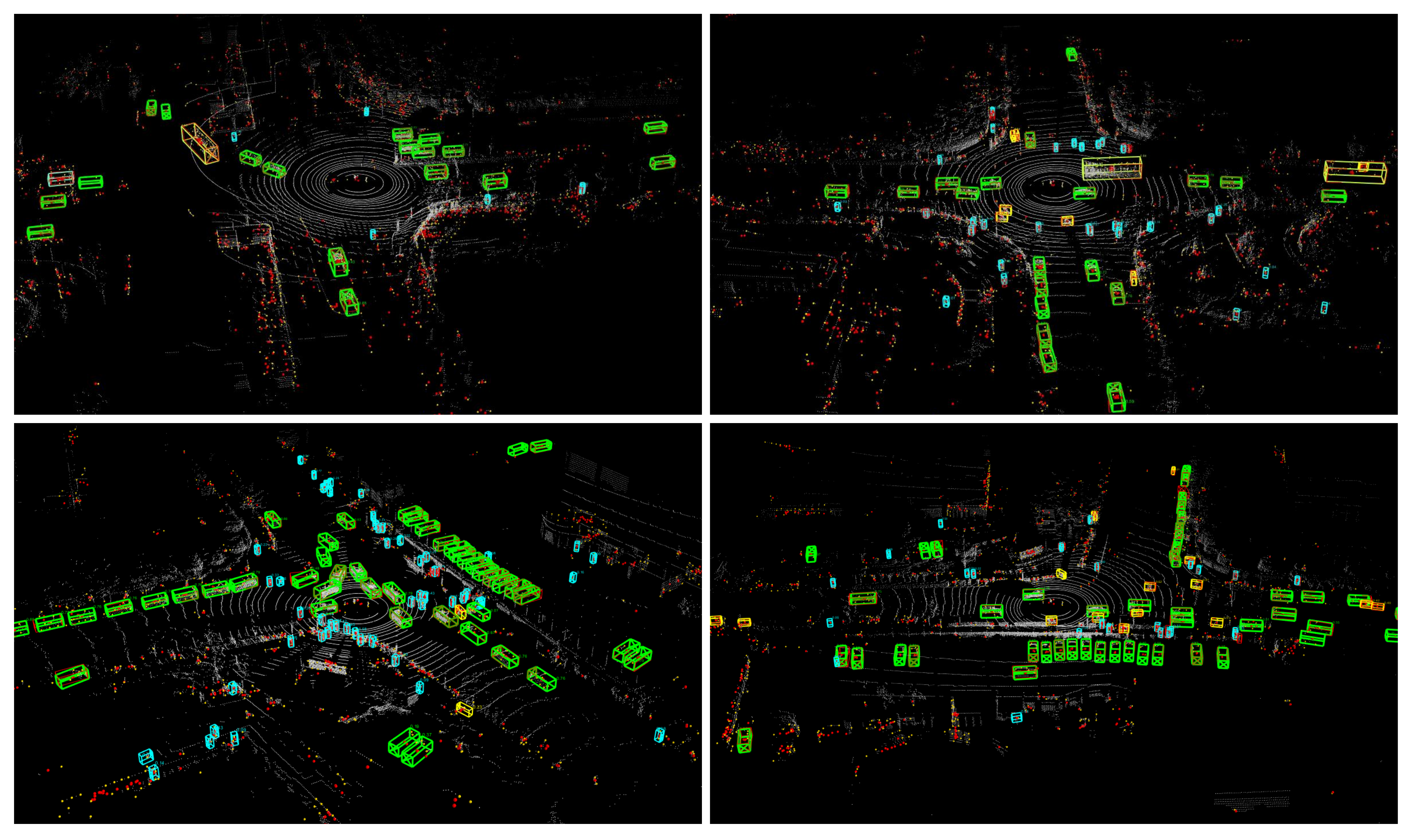}
    \caption{\yf{Extra qualitative results achieved by our \nickname{} on the \textit{val} set of the ONCE Dataset. Here We demonstrate our detection results on some challenging scenes. Note that, the ground-truth bounding boxes are shown in red, and the predicted bounding boxes are shown in green for \textit{vehicle}, cyan for \textit{pedestrian}, and yellow for \textit{cyclist}. Best viewed in color.}}
   \label{fig:visual_once_val}
\end{figure*}

\smallskip\noindent\textbf{(8) Additional qualitative detection results on the large-scale datasets.} Here, we present extra qualitative detection results achieved by our \nickname{} on two large-scale datasets with challenging panoramic scenarios. Figure~\ref{fig:visual_waymo_val} and Figure~\ref{fig:visual_once_val} illustrate the detection results on the validation set of Waymo and ONCE Dataset respectively. It can be seen that our \nickname{} can also achieve promising detection performance in challenging and complex 3D scenes. 

% \section{Limitation and Future Work}
% \smallskip\noindent\textbf{Limitation.} Although the proposed \nickname{} can achieve remarkable efficiency in object detection of large-scale LiDAR points clouds, it also has limitations. \textit{e.g.,} the instance-aware sampling relies on the semantic prediction of each point, which is susceptible to class imbalances distribution. For future work, we will further explore advanced techniques to alleviate the imbalanced issue.

\section{Potential Negative Societal Impact}
\qy{In this paper, we proposed an efficient point-based solution capable of achieving promising low-cost objects detection in autonomous driving scenarios. Our model is trained and evaluated totally based on open-sourced datasets, and there is no known potential negative impact on society.}

\section{Video Illustration}
We provide a video demo illustrating the detection performance of our IA-SSD in 3D point clouds, which can be viewed at \url{https://youtu.be/3jP2o9KXunA}.

\clearpage

\end{document}